\documentclass[letterpaper, 10 pt, conference]{ieeeconf}  
\IEEEoverridecommandlockouts                              
\usepackage{epsfig} 
\usepackage{times} 
\usepackage{amsmath} 
\usepackage{hyperref}
\usepackage[capitalise]{cleveref}
\usepackage{amssymb}  
\usepackage{balance} 
\usepackage{subcaption}
\usepackage{caption}
\usepackage[per-mode=symbol]{siunitx}
\usepackage{lipsum} 
\usepackage{enumerate} 
\usepackage{gensymb} 
\usepackage{graphicx}
\usepackage{adjustbox}
\usepackage{booktabs}
\usepackage{multirow}
\usepackage[dvipsnames]{xcolor}
\usepackage{stfloats} 
\usepackage{tikz}
\usepackage{algorithm}
\usepackage{algpseudocode}
\usepackage{svg}
\usetikzlibrary{patterns} 
\usepackage{pgfplots}

\usepackage{xcolor}

\usepackage{pgfplotstable}

\usepackage{tikz}
\usetikzlibrary{external,positioning}
\usepgfplotslibrary{external}
\usepackage{soul}

\makeatletter
\let\NAT@parse\undefined
\makeatother

\usepackage[numbers,sort&compress]{natbib}

\title{\LARGE \bf Rapid Quadrotor Navigation in Diverse Environments using an Onboard Depth Camera}
\author{Jonathan Lee, Abhishek Rathod, Kshitij Goel, John Stecklein, and Wennie Tabib}

\begin{document}
\bstctlcite{IEEEexample:BSTcontrol}
{
\twocolumn[{%
\begin{@twocolumnfalse}
\maketitle
\thispagestyle{empty}
\pagestyle{empty}
\vspace{-0.75cm}
\begin{figure}[H]
  \begin{minipage}{\textwidth}
    \centering
    \subcaptionbox{Cave\label{sfig:cave}}{\includegraphics[height=4.8cm,trim=700 0 0 0,clip]{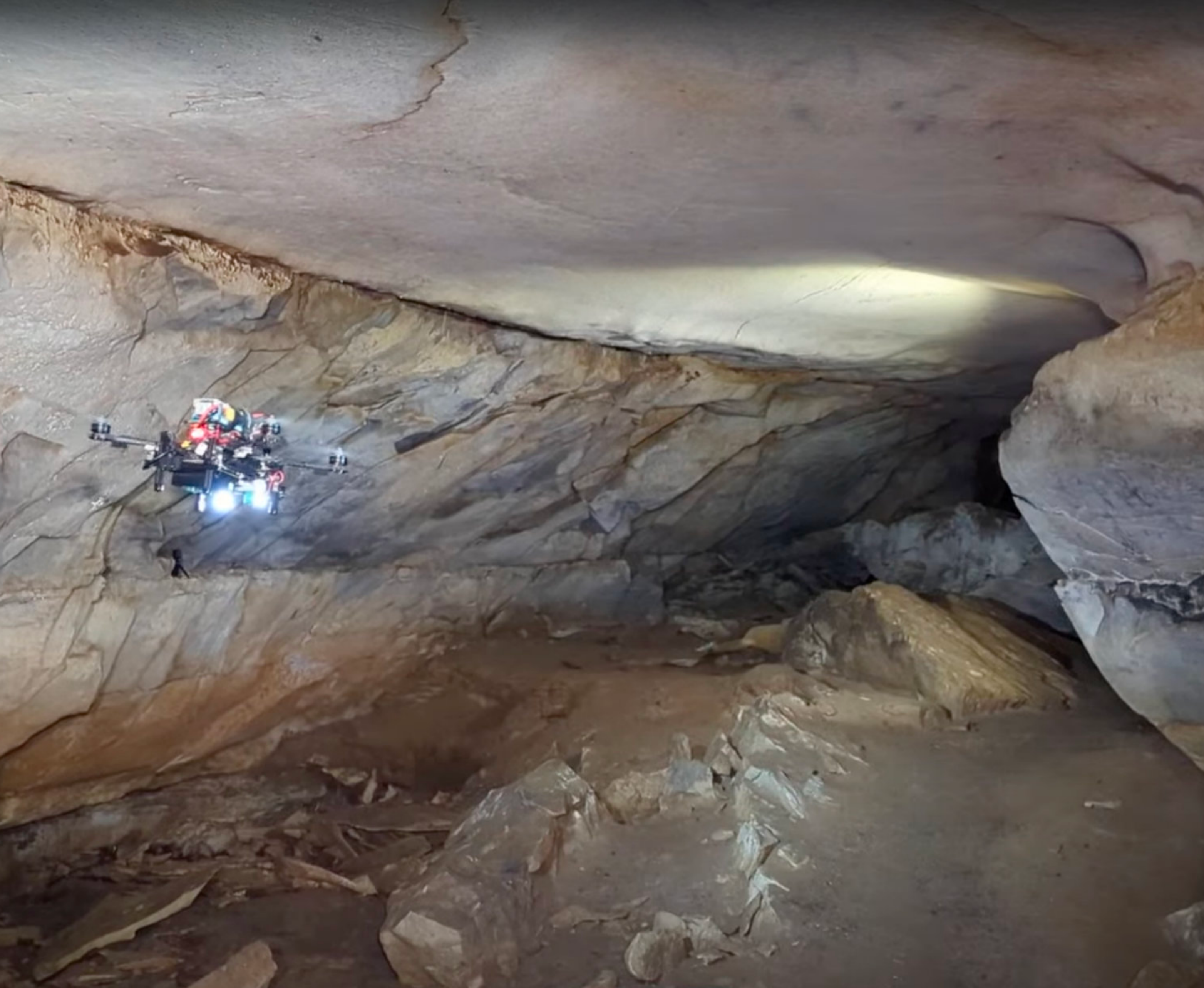}}\hfill
    \subcaptionbox{Industrial tunnel\label{sfig:industry}}{\includegraphics[height=4.8cm,trim=0 0 600 200,clip]{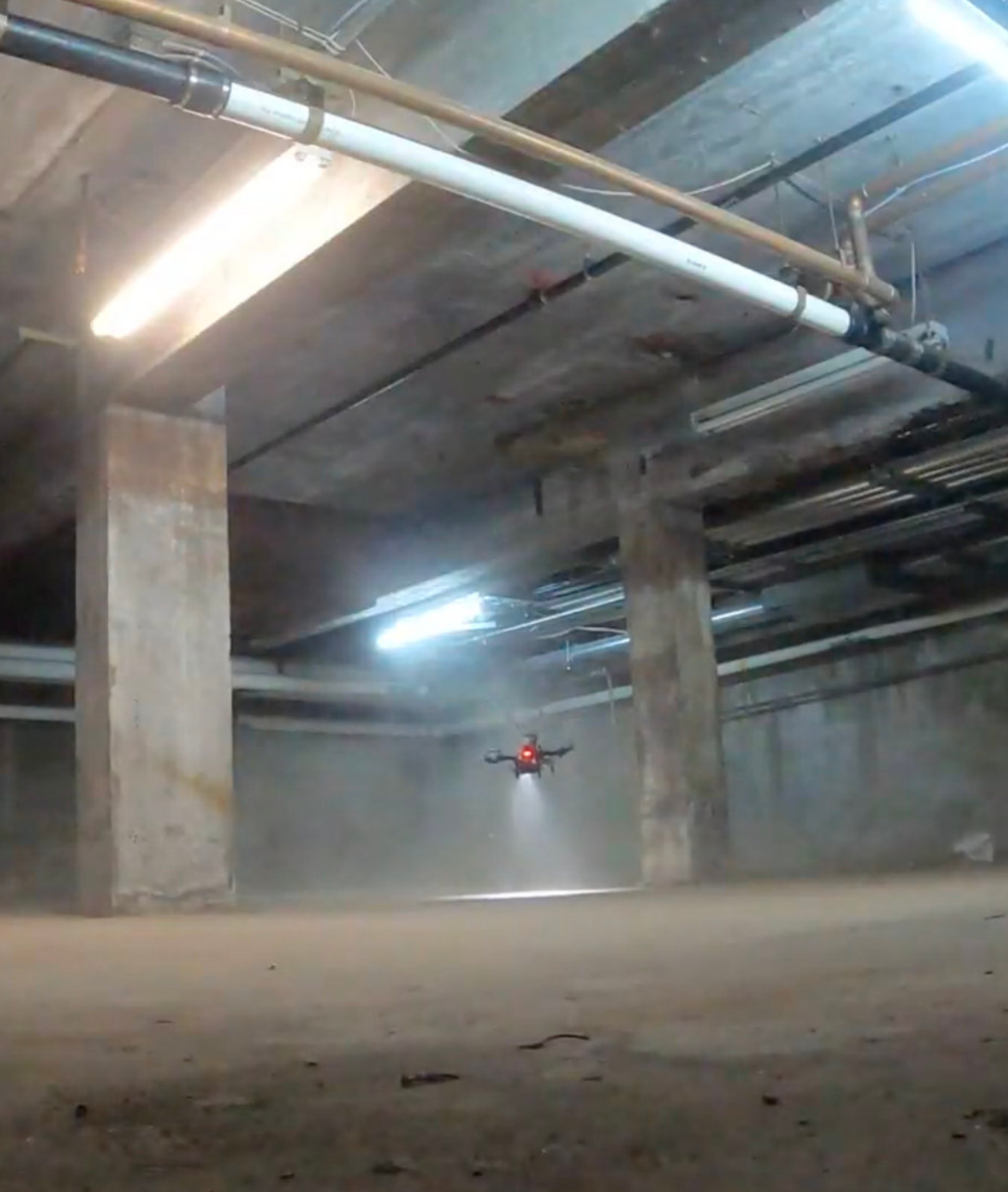}}\hfill
    \subcaptionbox{Forest\label{sfig:forest-glory}}{\includegraphics[height=4.8cm,trim=0 0 500 0,clip]{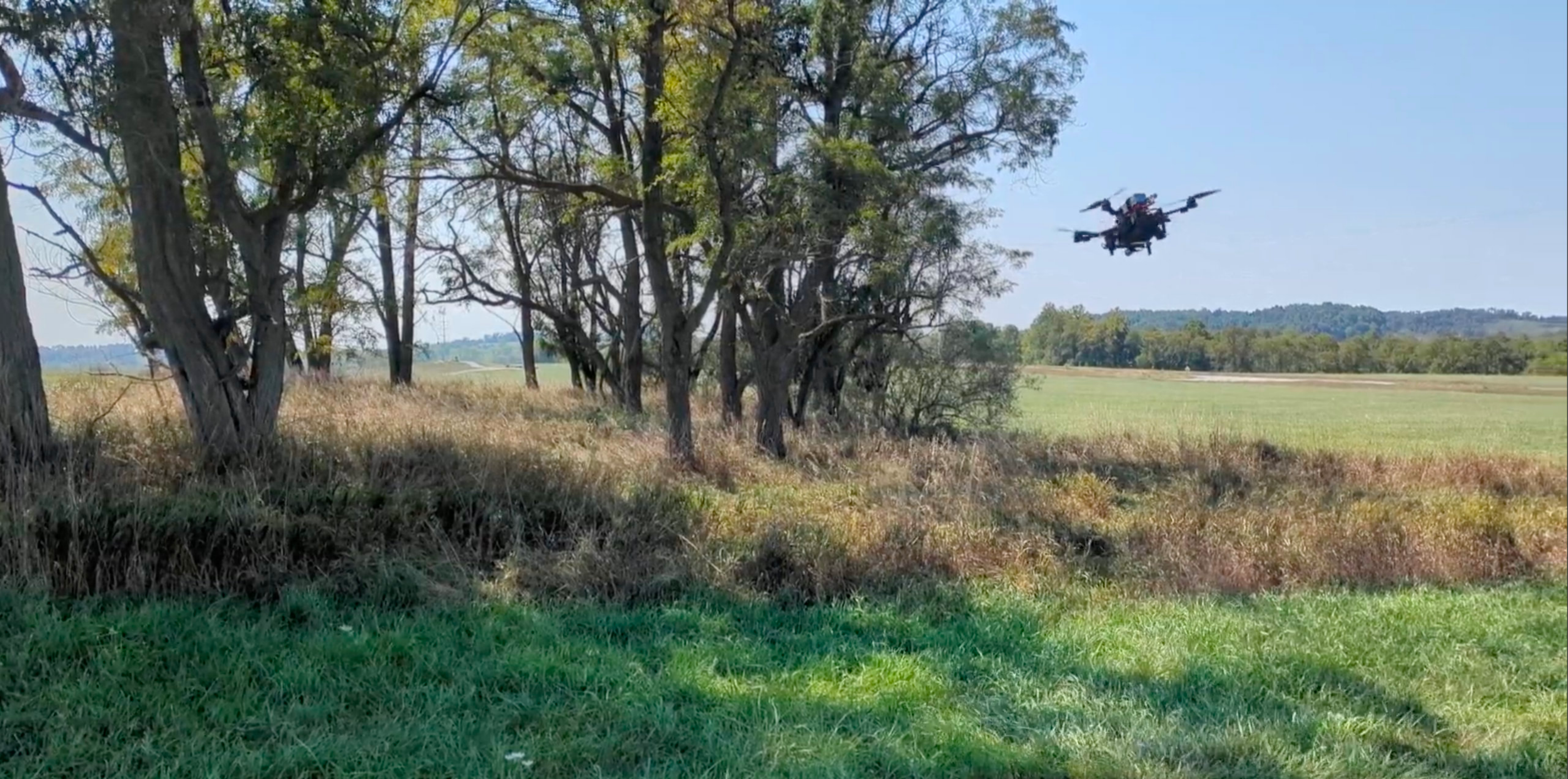}}\\
    \caption[\label{fig:gloryshot}]{(\protect\subref{sfig:cave}) Tight
        spaces,~(\protect\subref{sfig:industry}) high amounts of particulate
        matter such as dust and~(\protect\subref{sfig:forest-glory}) thin
        obstacles such as branches are hazards found in
        search and rescue environments. To aid search teams,
        autonomous aerial systems must rapidly navigate these dangers without
        posing additional risks for rescuers or victims.
        This paper proposes a rapid quadrotor navigation system, which uses forward-arc motion
        primitives and a forward-facing depth camera to achieve speeds up
        to $\SI{6}{\meter\per\second}$ in cluttered environments. Safety
        is achieved by executing a safe stopping action when no feasible
        action is found. Experiments are conducted in diverse environments, including caves and forests.
        A video of these experiments may be found at \url{https://youtu.be/tk8vUot0gD4}}%
  \end{minipage}
\end{figure}
\end{@twocolumnfalse}
}]
{
  \renewcommand{\thefootnote}%
  {\fnsymbol{footnote}}
  \footnotetext[1]{This work was supported in part by the U.S. Army
    Research Office and the U.S. Army Futures Command under Contract
    No. W519TC-23-C-0031.\\ The authors are with the Robotics Institute,
    Carnegie Mellon University, Pittsburgh, PA 15213 USA.  (email:
    \{\texttt{jlee6,arathod2,kgoel1,jsteckle,wtabib}\}
    \texttt{@andrew.cmu.edu}.)}%
}%
}

\newcommand{\bodyframe}{\mathcal{B}}
\newcommand{\bodyframex}{\mathbf{x}^\mathcal{B}}
\newcommand{\bodyframey}{\mathbf{y}^\mathcal{B}}
\newcommand{\bodyframez}{\mathbf{z}^\mathcal{B}}

\begin{abstract}
  Search and rescue environments exhibit challenging 3D
geometry (e.g., confined spaces, rubble, and breakdown), which
necessitates agile and maneuverable aerial robotic systems.  Because
these systems are size, weight, and power (SWaP) constrained, rapid
navigation is essential for maximizing environment coverage.  Onboard
autonomy must be robust to prevent collisions, which may endanger rescuers and
victims. Prior works have developed high-speed navigation solutions
for autonomous aerial systems, but few have considered safety for
search and rescue applications. These works have also not demonstrated
their approaches in diverse environments.  We bridge this gap in the
state of the art by developing a reactive planner using forward-arc
motion primitives, which leverages a history of RGB-D observations to
safely maneuver in close proximity to obstacles.
At every planning round, a safe stopping action is scheduled,
which is executed if no feasible motion plan is found at the
next planning round.  The approach is evaluated in thousands of
simulations and deployed in diverse environments, including caves and
forests. The results demonstrate a 24\% increase
in success rate compared to state-of-the-art approaches.
\end{abstract}

\section{Introduction\label{sec:intro}}
Extreme search and rescue environments exhibit challenging 3D geometry (e.g.,
confined spaces, rubble, breakdown), which preclude rapid traversal by legged,
wheeled, and tracked robotic systems. Quadrotors are highly agile and
maneuverable, but are limited by battery capacity and flight time. Therefore,
these systems must rapidly navigate through the environment. More importantly,
onboard autonomous navigation must be robust as collisions endanger rescuers and
victims.  To achieve these goals in highly cluttered environments with narrow
gaps, it is advantageous to leverage smaller size robots equipped with a
lightweight, short-range depth camera as opposed to a heavy, long-range LiDAR.

In this work, a rapid quadrotor navigation methodology is proposed, which
leverages a forward-facing, limited field-of-view RGB-D camera to operate in
diverse environments (e.g., caves, sewers, forests, industrial tunnels) without
scene-specific parameter tuning or training.  A reactive planner is developed
using forward-arc motion primitives, which are differentiable up to jerk and
continuous up to snap to enable aggressive flight. Motion primitive selection is
based on a perception front-end that searches a history of RGB-D observations to
safely maneuver in close proximity to obstacles.

The contributions of this work are: (1) a reactive navigation framework that
uses a history of depth observations to evaluate a library of forward-arc motion
primitives; (2) a trajectory scheduling approach to prevent collisions; (3)
extensive photo-realistic simulations with 3150 trials in diverse cluttered
environments; and (4) hardware experiments covering \SI{571}{\meter} in
successful trials without collisions and achieving a maximum speed of
$\SI{6}{\meter\per\second}$.  The results demonstrate a $24\%$ higher success
rate compared to state-of-the-art approaches.

\section{Related Work\label{sec:related_work}}
This section reviews collision avoidance and planning methods for agile
navigation with RGB-D sensors in unknown environments.

Learning-based methods have recently been proposed for rapid
flight~\citep{loquercioLearningHighspeedFlight2021,zhangBackNewtonLaws2024};
however, these methods lack strict collision-free guarantees and may not
generalize to environments outside the training domain, which are necessary for
deployment in disaster response scenarios. Reactive (memoryless) methods, on the
other hand, act directly on instantaneous sensor data or maintain a short
spatio-temporal history.
RAPPIDS \citep{buckiRectangularPyramidPartitioning2020} generates trajectories
within the latest depth image and iteratively partitions free space into
rectangular pyramids for efficient collision detection; however, the planner is
prone to becoming trapped in cluttered environments due to the pyramidal
corridor constraints and lack of yaw control.
BiTE~\citep{viswanathanEfficientTrajectoryLibrary2020} accelerates collision
checks with a local occupancy grid by pre-computing a bitwise map and trading
off increased memory usage for a fixed path library. This library, however, is
not continuous for higher-order derivatives of velocity.
NanoMap \citep{florenceNanoMapFastUncertaintyAware2018} searches for the minimum
uncertainty view of a queried point in space by maintaining a short temporal
history of depth images and their relative poses.
\citet{jiMaplessPlannerRobustFast2021} improves upon NanoMap and constructs a
forward spanning tree to find a safe flight corridor for trajectory generation.

Our approach is most similar to
\citet{florenceIntegratedPerceptionControl2020}, which utilizes
NanoMap for collision queries. However, their trajectory
representation demands high control effort
and lacks guaranteed safe stopping behaviors.
In contrast, our approach always includes a safe stopping trajectory, ensuring
the robot will stop in known free space if no feasible action exists.

\section{Technical Approach\label{sec:methodology}}
\begin{figure}[!t]
    \centering
    \includegraphics[width=0.99\linewidth,trim=20 20 20 20,clip]{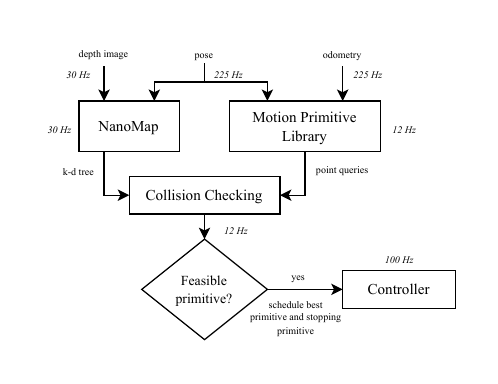}
    \caption{\label{fig:system_diagram}System diagram of the navigation algorithm. Given depth
    images and odometry, NanoMap~\citep{florenceIntegratedPerceptionControl2020} is used for collision avoidance
    and a library of forward-arc motion primitives is generated for motion planning. To
    maintain safety, collision-free trajectories are scheduled such that
    a feasible stopping action is always available within the known free space.}
    \vspace{-0.5cm}
\end{figure}

\begin{figure}[!t]
    \centering
    \includegraphics[width=0.99\linewidth]{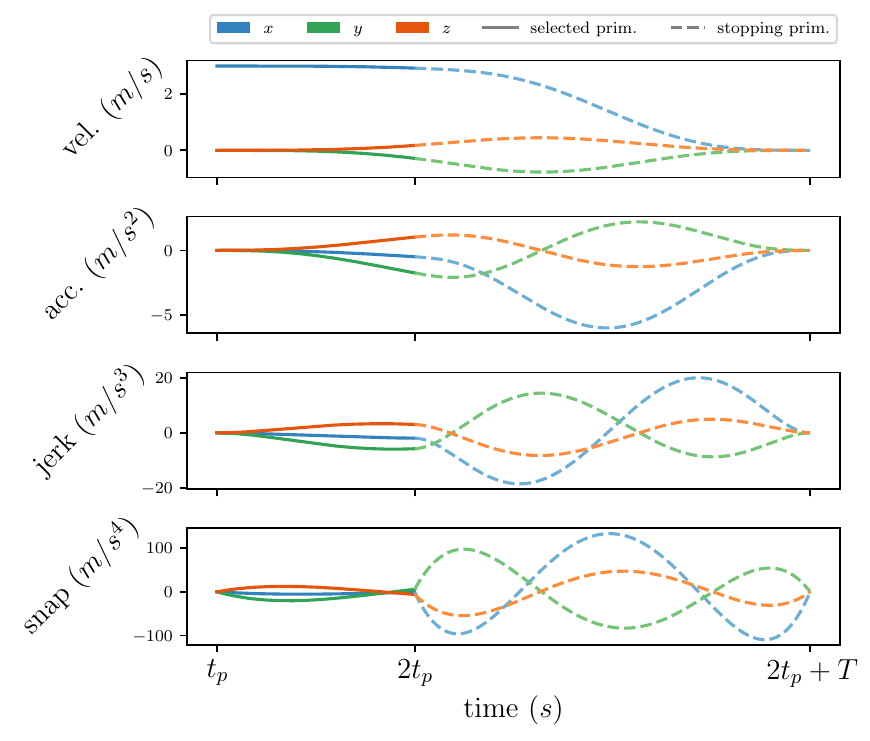}
    \caption{\label{fig:derivatives}
    Derivatives of the scheduled trajectory are continuous up to snap and smooth
    up to jerk. The planning strategy ensures the robot stops in a safe
    region.}
    \vspace{-0.5cm}
\end{figure}

This section details the perception and planning strategy for rapid
navigation. The high-level system architecture for the
approach is illustrated in~\cref{fig:system_diagram}. A
forward-arc motion primitive library is used to generate smooth
actions, which are evaluated for collisions. The perception
system maintains a 3D point cloud representation of the local
environment and searches over a history of observations. Of
the feasible motion primitives, the one with the lowest cost is
scheduled for execution. Additionally, a safe stopping primitive is
scheduled, which is executed if no feasible motion primitive is found
at the next planning round.

\subsection{Forward-Arc Motion Primitive Library}
Forward-arc motion primitives propagate the dynamics of a unicycle model for
time $T$ and are parameterized by a desired yaw rate and forward and
vertical velocities~\cite{yang2017framework}. This action formulation is differentiable
up to jerk and continuous up to snap (see~\cref{fig:derivatives}), which is advantageous for aggressive
multirotor flight where large angular velocities and accelerations are directly
related to the jerk and snap of the reference
position~\cite{spitzer2020fast,yang2019online}.

\begin{figure}[!t]
    \centering
    \includegraphics[width=0.99\linewidth,trim=10 35 30 25,clip]{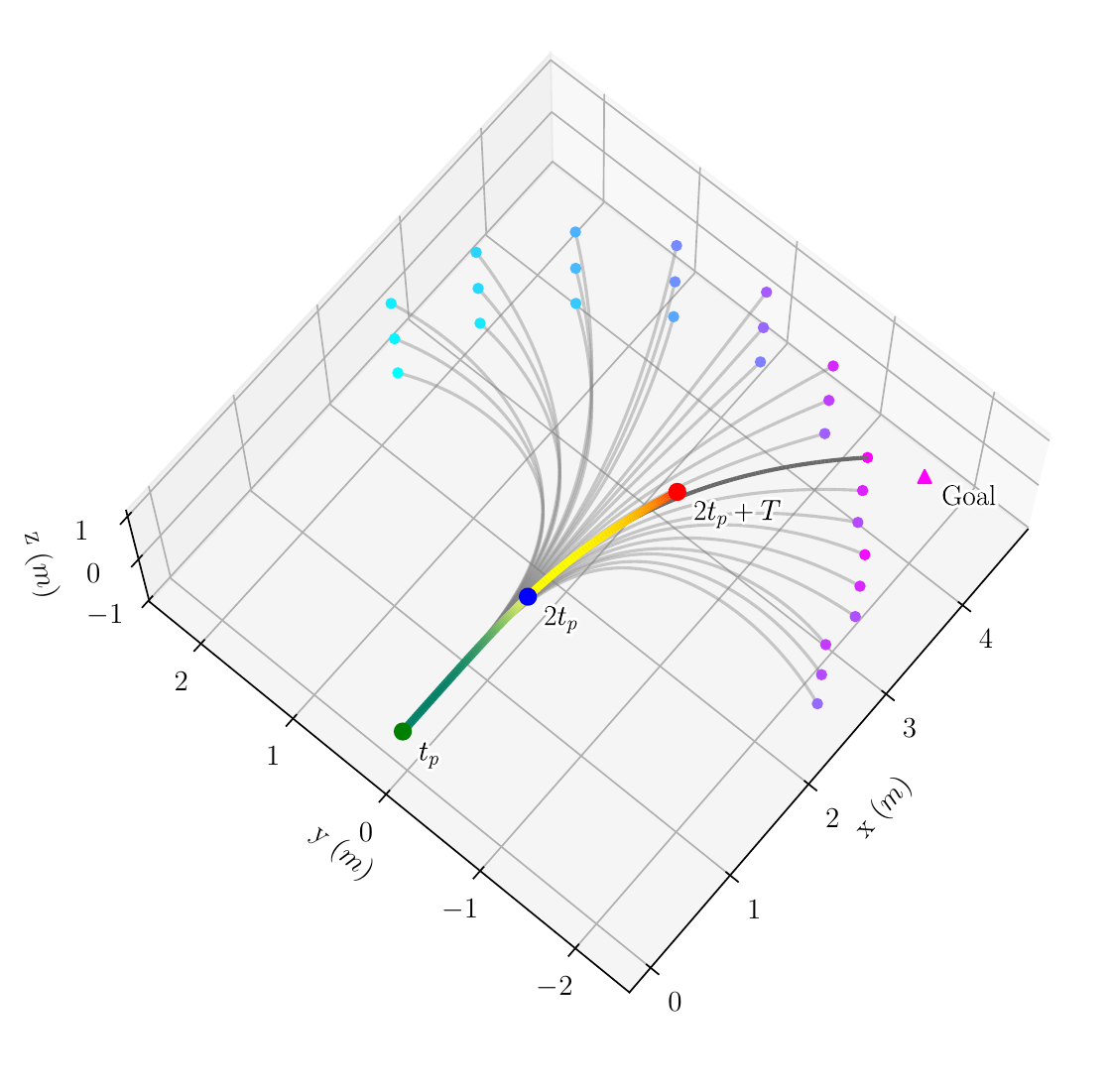}
    \caption{\label{fig:mpl}
    Illustrative example of trajectory scheduling with the motion primitive library.
    The motion primitives in the library are shown in gray.
    The cost of each primitive is evaluated by computing the Euclidean distance between the
    endpoint and goal (shown as a pink triangle). The endpoints
    are shown as dots colored from purple to blue, where more pink indicates closer to the
    goal.  Motion primitives that are in collision are pruned. The primitive with
    lowest cost (shown in dark gray) is selected for execution. The selected primitive
    segment is scheduled from times $[t_p, 2t_p)$ and the stopping primitive is
    scheduled from $[2t_p, 2t_p+T)$.}
    \vspace{-0.5cm}
\end{figure}

Let $\bodyframe$ represent the body frame of the multirotor. The
position and heading of the vehicle is represented as $\pmb{\xi}_t
= \begin{bmatrix} x & y & z & \theta\end{bmatrix}^{\top}$. The linear
velocities are expressed in the body frame as $\begin{bmatrix}
v_{x,t}^{\bodyframe} & v_{z,t}^{\bodyframe} \end{bmatrix}$. The angular
velocity, $\omega_{z,t}^{\bodyframe}$ is expressed about the
$\bodyframez$ axis. The equations of the motion primitives are given
by the solutions to the unicycle
model~\cite{yang2019online,tabib2021autonomous}:
\begin{align}
  \pmb{\xi}_{t+T} &= \pmb{\xi}_t + \begin{bmatrix}
    \frac{v_{x,t}^{\bodyframe}}{\omega_{z,t}^{\bodyframe}}(\sin(\omega_{z,t}^{\bodyframe}T + \theta_t) - \sin(\theta_t))\\
    \frac{v_{x,t}^{\bodyframe}}{\omega_{z,t}^{\bodyframe}}(\cos(\omega_{z,t}^{\bodyframe}T) - \cos(\omega_{z,t}^{\bodyframe}T + \theta_t))\\
    v_{z,t}^{\bodyframe}T\\
    \omega_{z,t}^{\bodyframe}T \end{bmatrix}.
\end{align}
The motion primitive, $\gamma$, is parameterized by
$\mathbf{a}_t = \{v^{\bodyframe}_{x,t}, v^{\bodyframe}_{z,t},
\omega_t^{\bodyframe}\}$ and duration $T$. $v^{\bodyframe}_{x,t}$ is
fixed by the user. $v^{\bodyframe}_{z,t} \in \mathcal{V}_{z}$ and
$\omega_t^{\bodyframe} \in \Omega$ are varied according to a
user-specified discretization. The motion primitive library, $\Gamma$,
is generated as the set of motion primitives created by varying
$v^{\bodyframe}_{z,t}$ and $\omega_t^{\bodyframe}$ (see~\cref{fig:mpl} for
an illustration of the library).

Motion primitives are scheduled at a fixed planning rate as shown in~\cref{fig:mpl,fig:derivatives}.
The selected primitive is executed from
$[t_p, 2t_p)$. The stopping primitive is scheduled from
$[2t_p, 2t_p +T)$ so that if the next planning round fails to find a feasible
motion primitive, the robot stops in free space.

\subsection{Local 3D Perception and Collision Checking\label{subsec:3d_perception}}
Nanomap~\citep{florenceNanoMapFastUncertaintyAware2018} is used
for high-rate proximity queries for collision avoidance. We provide a
brief overview of the query search algorithm, which uses $k$-nearest
neighbor queries over a short temporal history of depth measurements.

NanoMap maintains a chain of edge-vertex pairs where each edge
contains the relative transform
$T_{\mathcal{S}_{i-1}}^{\mathcal{S}_i}$ between consecutive sensor
frames and each vertex contains the ordered point cloud and corresponding
k-d tree.  The motion primitive representing the position of the
vehicle, $\gamma^{(0)}$, is sampled in time using a fixed $\Delta
t$ up to the trajectory duration, $T$, to generate query points,
$\mathbf{p}_{\text{query}}^\mathcal{B} \in \mathbb{R}^3$.
\begin{align}
\mathbf{p}_{\text{query}}^\mathcal{B} &= \gamma^{(0)}(t).
\end{align}
Query points sampled from the motion plan
 are iteratively
transformed into previous sensor frames until a view containing the
query point is found. The transformation to the $i$-th coordinate
frame is obtained by
\begin{align}
    \mathbf{p}_{\text{query}}^{\mathcal{S}_i} = \prod_{j=1}^{i}\left[ T_{\mathcal{S}_{j-1}}^{\mathcal{S}_j}\right] T_\mathcal{B}^{\mathcal{S}_0} \mathbf{p}_{\text{query}}^{\mathcal{B}}
  \label{eq:nanomap}
\end{align}
Once a query is determined to be in view, the ordered point cloud is used
to determine whether the point is within free space, after which the
k-d tree is used to evaluate the $k$-nearest neighbors.  Note that
in~\cref{eq:nanomap}, the uncertainty propagation of the query point
is disabled. This modification is made because we leverage a more
accurate optimization-based visual inertial navigation system
(detailed in following sections) for state estimation compared to what
is used in~\cite{florenceIntegratedPerceptionControl2020}.

For each query position, $\mathbf{p}_{\text{query}}^{\bodyframe}$, the distance,
$d$, is found between $\mathbf{p}_{\text{query}}^{\bodyframe}$ and the nearest
neighbor in the depth image. For a user-specified collision radius
$r_{\text{coll}}$, the primitive is marked infeasible if $d < r_{\text{coll}}$.
Feasible primitives are scored using the Euclidean distance to the goal position, $g$,
from the end of the primitive, $\texttt{Cost}(\gamma, g) =
\lVert g - \gamma^{(0)}(T) \rVert_2$. The primitive that minimizes the cost
function is scheduled.~\Cref{fig:system_diagram} shows an overview of the final algorithm
along with operating frequencies.

\begin{figure}[h!]
  \subcaptionbox{Cave\label{sfig:cave1}}{\includegraphics[width=0.329\linewidth, height=0.32\linewidth]{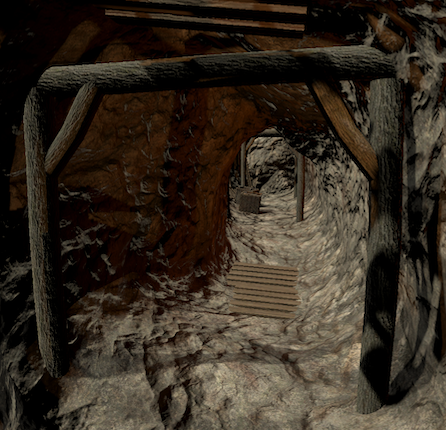}}\hfill
  \subcaptionbox{Industrial area\label{sfig:industry1}}{\includegraphics[width=0.329\linewidth, height=0.32\linewidth]{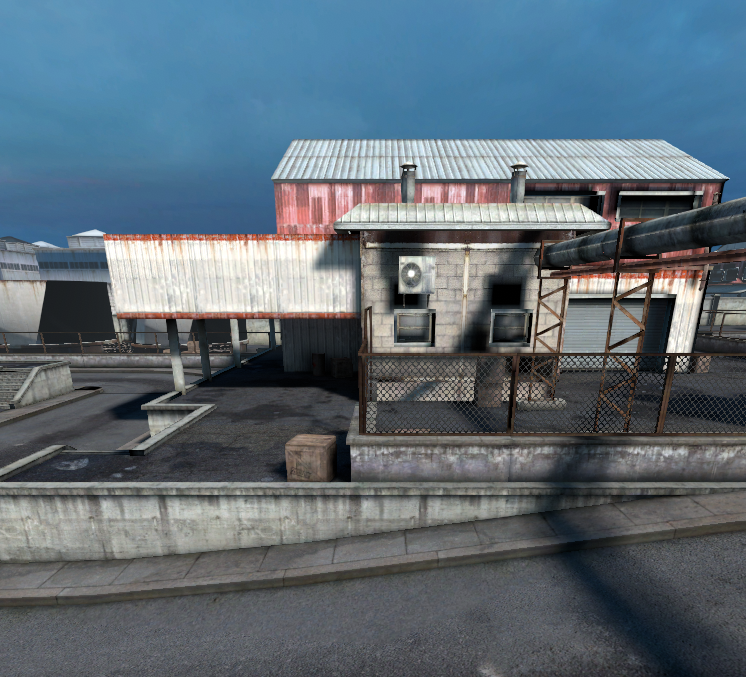}}\hfill
  \subcaptionbox{Sewer system\label{sfig:sewer5}}{\includegraphics[width=0.329\linewidth, height=0.32\linewidth]{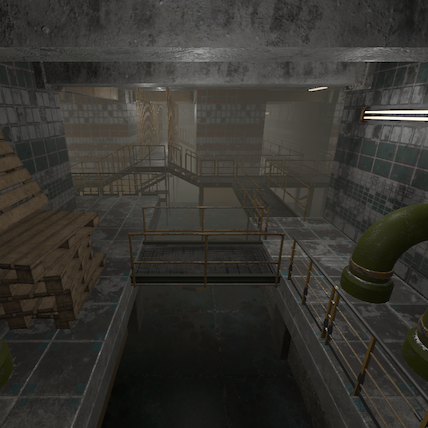}}%
  \caption{\label{fig:unity_scenes}
    (\protect\subref{sfig:cave1})--(\protect\subref{sfig:sewer5})
    illustrate a subset of the simulation environments used to validate
    the proposed approach.}
\end{figure}

\section{Results\label{sec:results}}
\begin{figure*}[!t]
    \centering
    \includegraphics[width=0.99\linewidth]{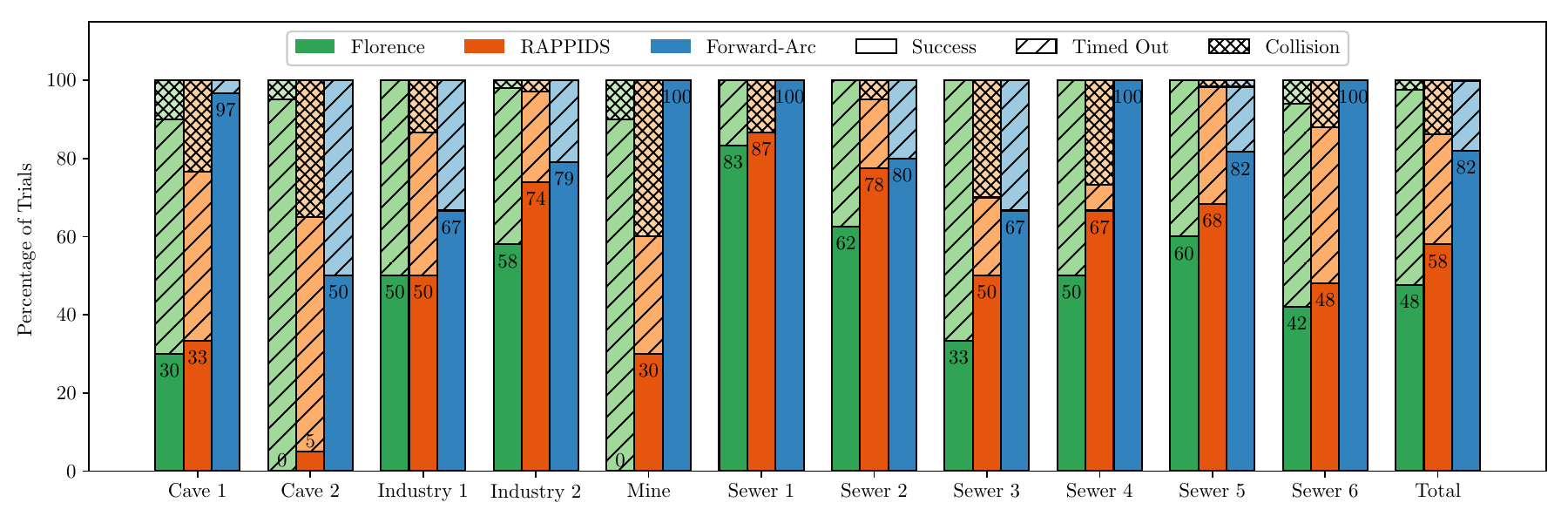}
    \caption{\label{fig:static_success_rate}Planner success rate and failure
    modes across the environments detailed in~\cref{table:unity_scenes}. The
    proposed method (\emph{Forward-Arc}) achieves the highest success rate and
    lowest collision rate compared to the baseline reactive planners
    \citep{florenceIntegratedPerceptionControl2020,
    buckiRectangularPyramidPartitioning2020}. A total of 1350 trials are
    run (450 for each approach).}
    \vspace{-0.5cm}
\end{figure*}

\begin{figure}[h]
    \centering
    \includegraphics[width=0.6\linewidth]{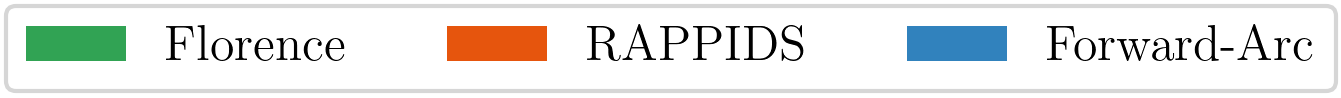}%
    \vspace{0.1cm}
    \includegraphics[width=0.99\linewidth,trim=600 200 250 200,clip]{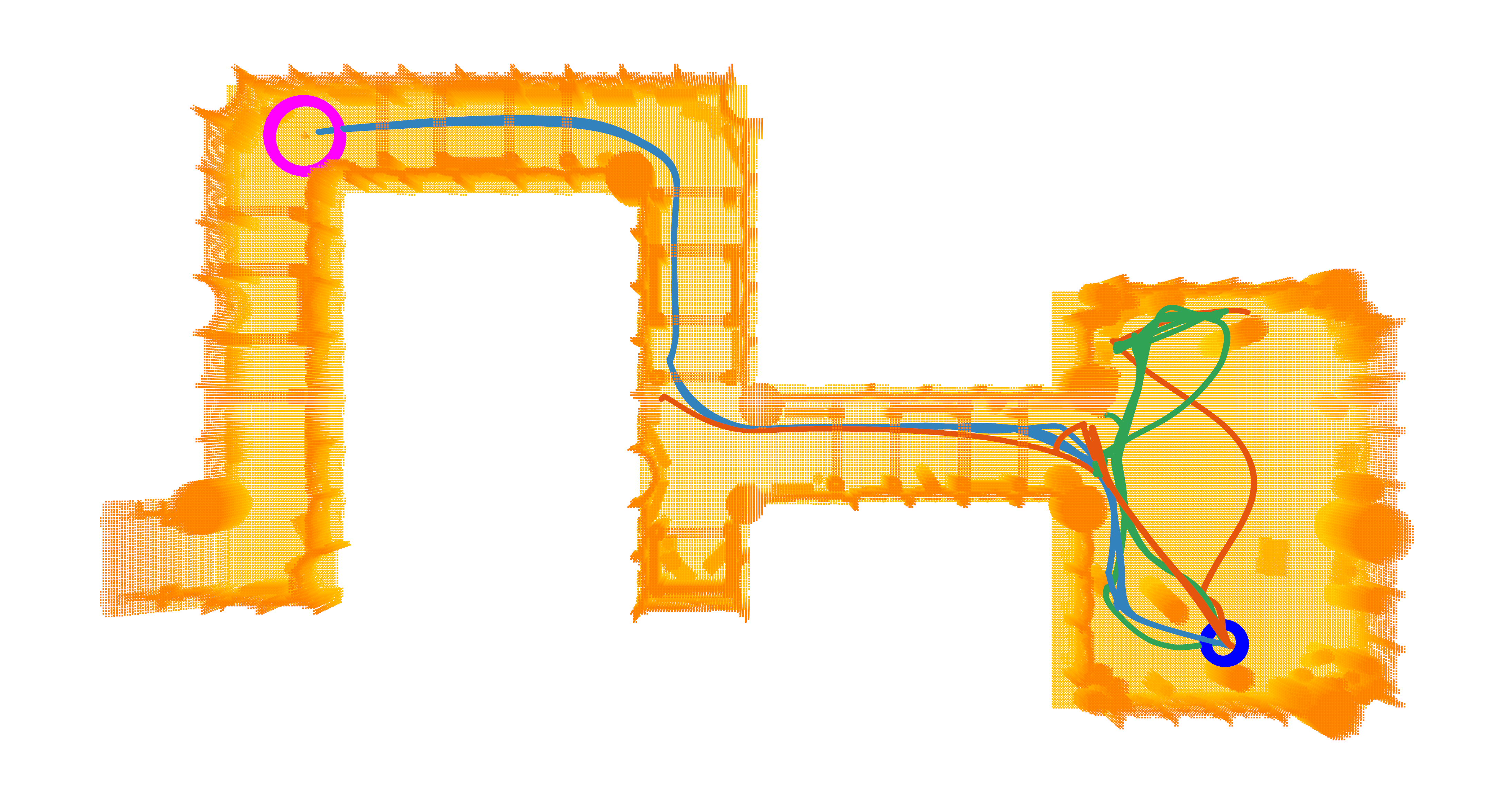}%
    \caption{Cross-section of the
      mine environment with overlays of 5 trajectories for each planning approach. The
      start and goal locations are indicated in blue and magenta, respectively.
      \emph{Forward-Arc} reaches the goal in all 5 trials. \emph{RAPPIDS}
      collides in 4 trials and times out in 1 trial. \emph{Florence}
      collides in 1 trials and times out in 4
      trials.\label{fig:trajectories}
    }
    \vspace{-0.25cm}
\end{figure}

The approach is evaluated in simulation
(see~\cref{ssec:disaster_sim,ssec:obstacle_density}) and with hardware
experiments (see~\cref{ssec:hardware_results}).  In simulation evaluations, the proposed
approach is compared against two reactive vision-based methods, RAPPIDS
\citep{buckiRectangularPyramidPartitioning2020}\footnote{\url{https://github.com/nlbucki/RAPPIDS}}
and
\citet{florenceIntegratedPerceptionControl2020}\footnote{\url{https://github.com/peteflorence/nanomap_ros}},
which demonstrate local planning in unknown environments.  For the
rest of this section, we will refer to these approaches as
\emph{RAPPIDS} and \emph{Florence}, respectively, and to the proposed approach
as \emph{Forward-Arc}.
We disable the ``dolphin'' oscillations
of~\cite{florenceIntegratedPerceptionControl2020}, which were used to
improve state estimation, because this is perfectly known in
simulation. Primitives that modulate z-height were found to cause
crashes due to altitude drift, so these were also removed. To set the
desired speed for \emph{RAPPIDS} we adjust the allocated trajectory
duration such that the planner achieves this speed in an open
field. We also use a velocity tracking yaw angle to enable navigation
in unstructured environments and generalize beyond the corridor
environments tested in
\citep{buckiRectangularPyramidPartitioning2020}.

Two simulation evaluations are conducted. First, diverse environments
representative of disaster scenarios are used to evaluate the success
rate of reaching a goal within a specified time
(detailed in~\cref{ssec:disaster_sim}).  Second, simulations that vary
the obstacle density (e.g. forest-like environments) are detailed
in~\cref{ssec:obstacle_density}. In addition to the success rates,
this second set of simulation experiments evaluates
the approaches in terms of flight time, path length, and control
effort. The Flightmare simulator~\cite{song2021flightmare} is used to generate
photorealistic images and depth maps at~\SI{30}{\hertz}. Simulations
are run on a desktop computer with the planner running on a single thread of an Intel i9-14900K CPU.
An NVIDIA RTX 4090 GPU is leveraged for photorealistic rendering within Flightmare.

\raggedbottom 
\begin{table}[h!]
\centering
\caption{Unity Custom Environments\label{table:unity_scenes}}
\begin{tabular}{lrrl}
\toprule
Scene & \# Goals & Trials & Description \\
\midrule
Cave 1 & 6 & 90 & Natural, Corridors \\
Cave 2 & 8 & 120 & Natural, Caverns \\
Industry 1 & 6 & 90 & Man-made, Buildings \\
Industry 2 & 20 & 300 & Man-made, Factory \\
Mine & 2 & 30 & Man-made, Corridor \\
Sewer 1 & 6 & 90 & Man-made, Rooms \\
Sewer 2 & 8 & 120 & Man-made, Rooms \\
Sewer 3 & 6 & 90 & Man-made, Rooms \\
Sewer 4 & 6 & 90 & Man-made, Tunnels \\
Sewer 5 & 12 & 180 & Man-made, Rooms \\
Sewer 6 & 10 & 150 & Man-made, Tunnels \\
\bottomrule
\end{tabular}
\end{table}

\subsection{Simulated Disaster Scenarios\label{ssec:disaster_sim}}

Simulations are conducted in caves, a mine, abandoned industrial
environments, and a sewer system (\cref{fig:unity_scenes}). Multiple start and goal locations are
specified to evaluate the robustness of each approach. For each pair
of start and goal locations each planner is run 5 times for a total
of 1350 trials (see \cref{table:unity_scenes}). A trial is considered successful if the robot reaches
the goal within a time limit. The sensing range is set to
\SI{10}{\meter} for all planning approaches.

The planner performance is evaluated by measuring
the success rate and failure modes (collisions and
timeouts). The results are reported in
\cref{fig:static_success_rate}. \emph{Forward-Arc}
outperforms the baseline methods in terms of success rate (82\%) and
has the fewest collisions (1 out of 450 trials).  In
contrast, we find that many \emph{Florence} collisions are caused by
aggressive emergency stopping maneuvers, which results in a
loss of stability. \Cref{fig:trajectories} provides a representative
figure for the performance of each of the approaches.
\emph{RAPPIDS} is unable to make sharp turns due to narrow pyramidal
flight corridor constraints. \emph{Florence}  is unable to enter the passageway
due to its conservative collision probability and uncertainty propagation
estimates. \emph{Forward-Arc} consistently navigates to the objective,
demonstrating robust and safe operation in diverse environmental conditions.

\subsection{Obstacle Density Experiments\label{ssec:obstacle_density}}%

\begin{figure}[t]
    \centering
    \includegraphics[width=0.6\linewidth]{figures/simulation/planner_legend.png}%
    \vspace{0.1cm}
    \includegraphics[width=0.99\linewidth,trim=400 300 500 500,clip]{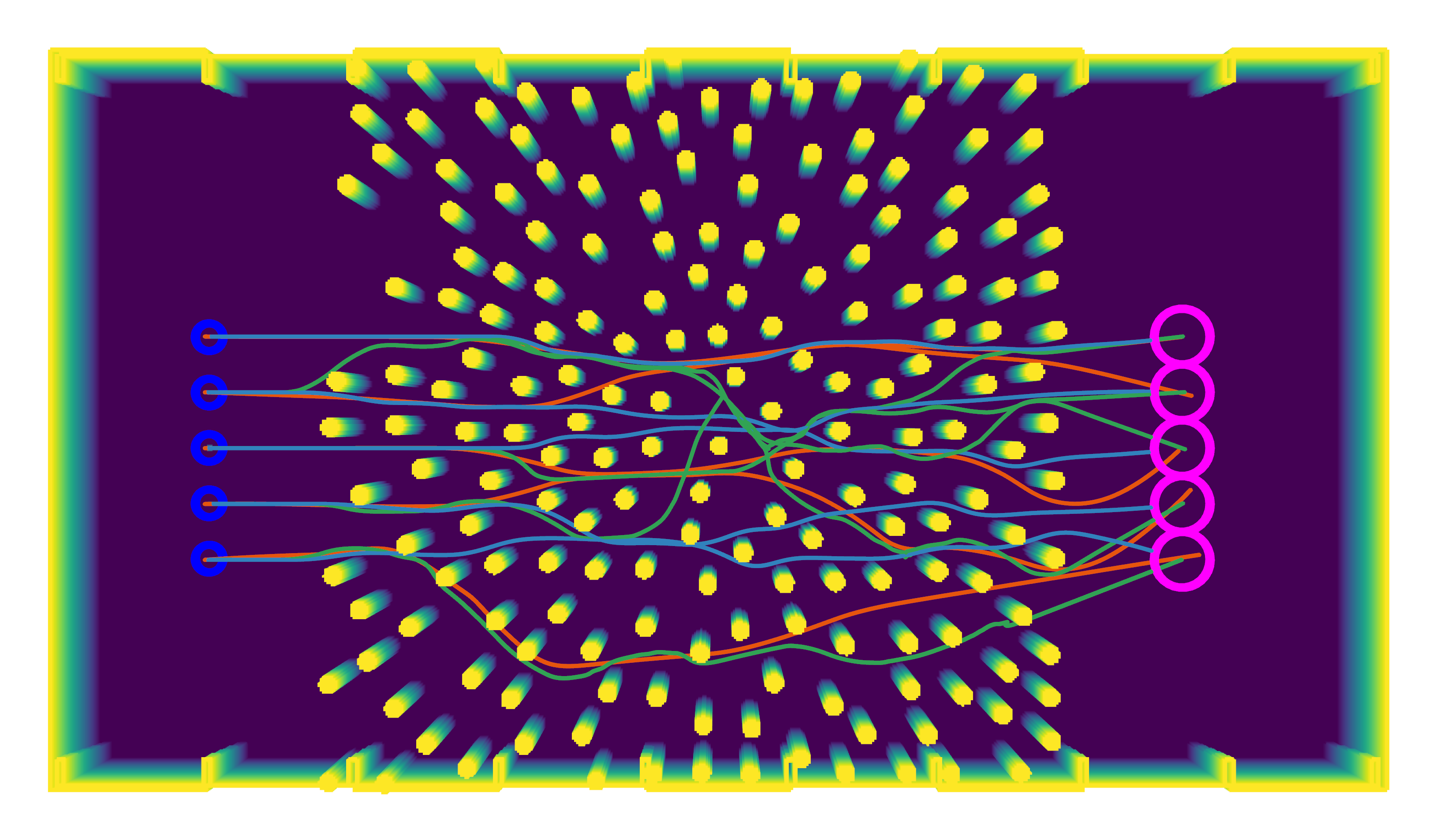}%
    \caption{Overlay of 5 trajectories for each planner on the ground truth
    point cloud ($v_{max}=3m/s$ and $\rho = 0.075 \text{ obstacles}/m^2$).
    \emph{Forward-Arc} takes the shortest path towards the goal compared to the
    baselines.\label{fig:obstacle_density_trajectories}}
    \vspace{-0.25cm}
\end{figure}

Next, we study the effect of varying obstacle densities and desired speeds on
planner performance. We design a forest-like environment using Poisson disk
sampling with a uniform density and cylindrical obstacles with a diameter of
0.75 m. Start and goal locations are positioned 70 meters apart with 10 evenly
spaced endpoints per random environment seed
(see \cref{fig:obstacle_density_trajectories} for example). We run a total of 1800 trials with
50 trials per planner at each combination in desired speed and obstacle
density.

\begin{figure*}[b!]
  \centering
  \begin{tabular}{cc}
  \adjustbox{valign=b}{\begin{tabular}{@{}c@{}}
      \includegraphics[width=0.43\linewidth]{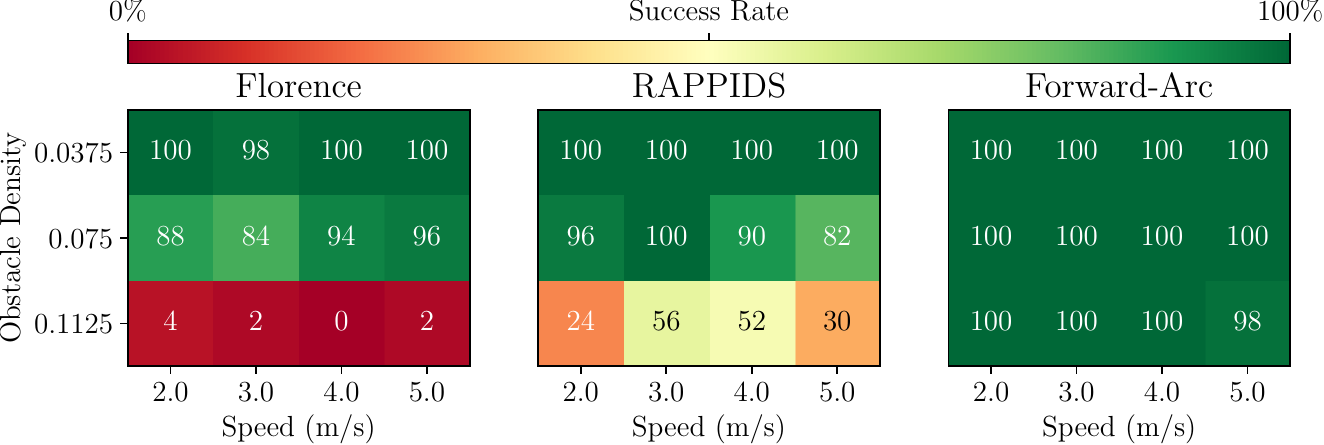}\\
      \subcaptionbox{\label{fig:density_speed_matrix}}{\includegraphics[width=0.43\linewidth]{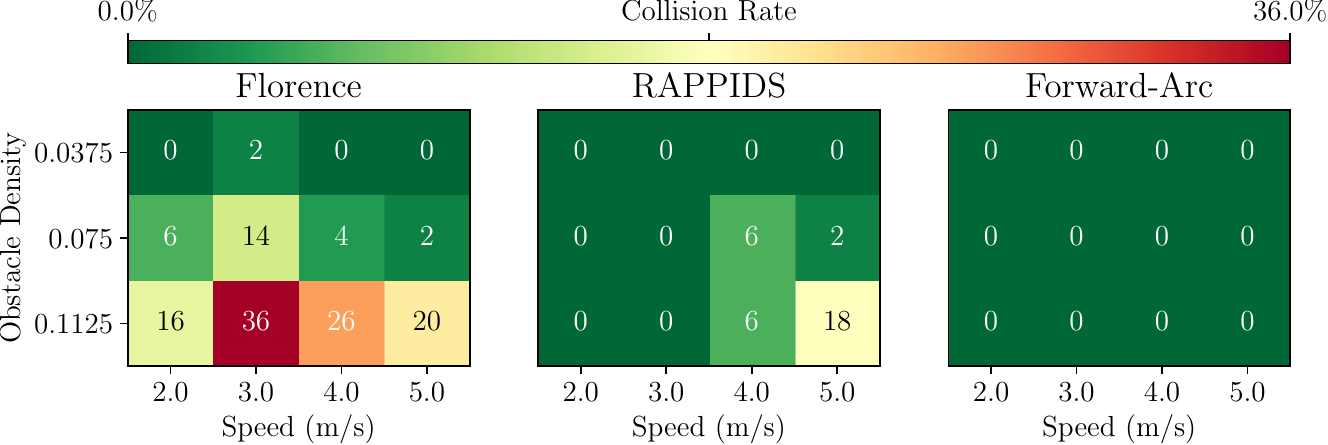}}
  \end{tabular}}
  &
      \adjustbox{valign=b}{\subcaptionbox{\label{fig:violin_box_plots}}{\includegraphics[width=0.53\linewidth,trim=12 12 10 5,clip]{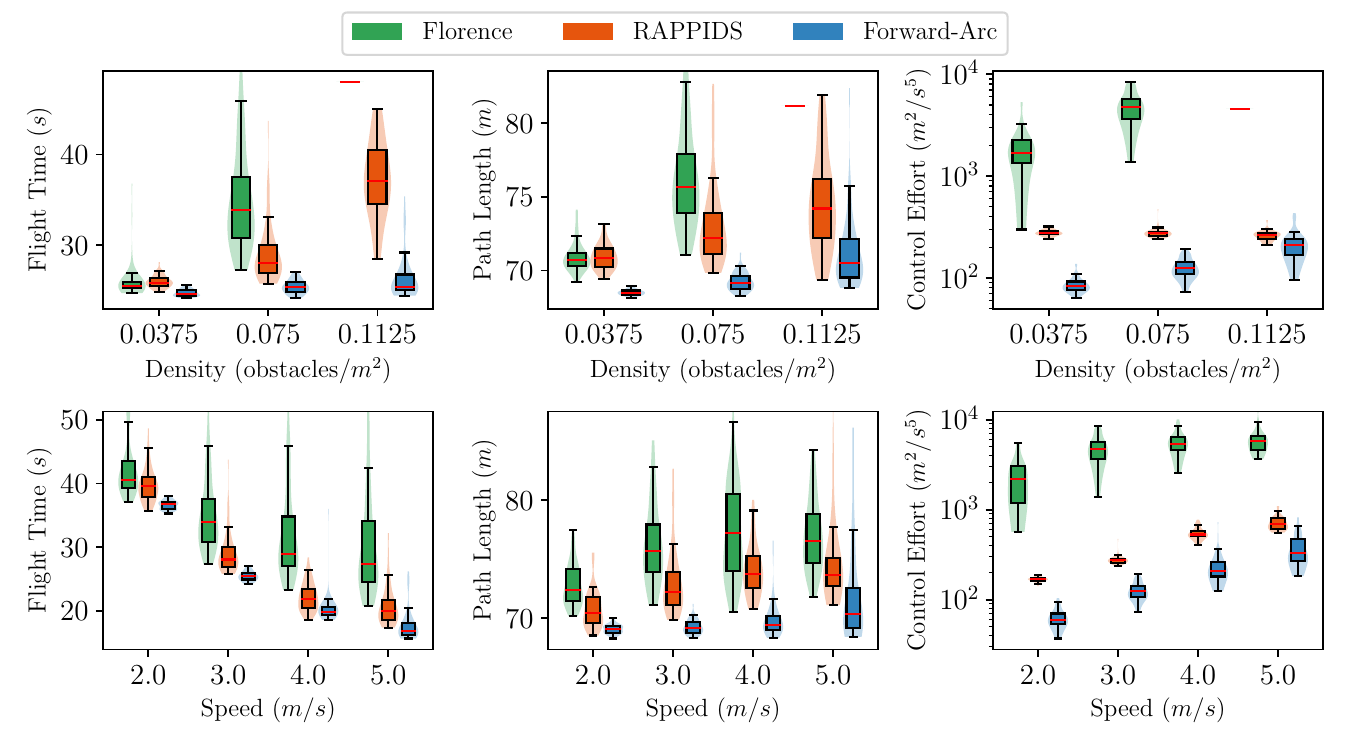}}}
  \end{tabular}
  \caption{\label{fig:obstacle_density}(\protect\subref{fig:density_speed_matrix}) success rate and
      collision rate matrices for simulations that vary obstacle
      densities and speeds. The \emph{Forward-Arc} approach has
      higher rates of success at higher obstacle densities and speeds
      compared to the baseline approaches.
      ~(\protect\subref{fig:violin_box_plots}) distribution of planning
      performance metrics from successful simulation trials across varying obstacle
      densities and speeds. Note that a log scale is used for the control effort
      plots. \emph{Forward-Arc} reaches the objective with lower average
      flight time, path length, and control effort metrics compared to the
      baseline approaches.}
\end{figure*}

In \cref{fig:density_speed_matrix}, \emph{Florence} struggles with high density
environments due to its aggressive stopping maneuvers and conservative collision
probability propagation. \emph{RAPPIDS} has an increased flight time at high
obstacle densities due to the lack of explicit yaw angle control leading it to
take longer paths (\cref{fig:obstacle_density_trajectories}). As shown in
\cref{fig:density_speed_matrix,fig:violin_box_plots}, \emph{Forward-Arc}
exhibits the lowest collision rate, flight time, path length, and control effort
(integral of jerk squared) and has the highest success rate amongst all
variations in obstacle densities and speeds.

\subsection{Hardware Experiments\label{ssec:hardware_results}}
\begin{figure}[h]
    \centering
    \includegraphics[width=0.59\linewidth]{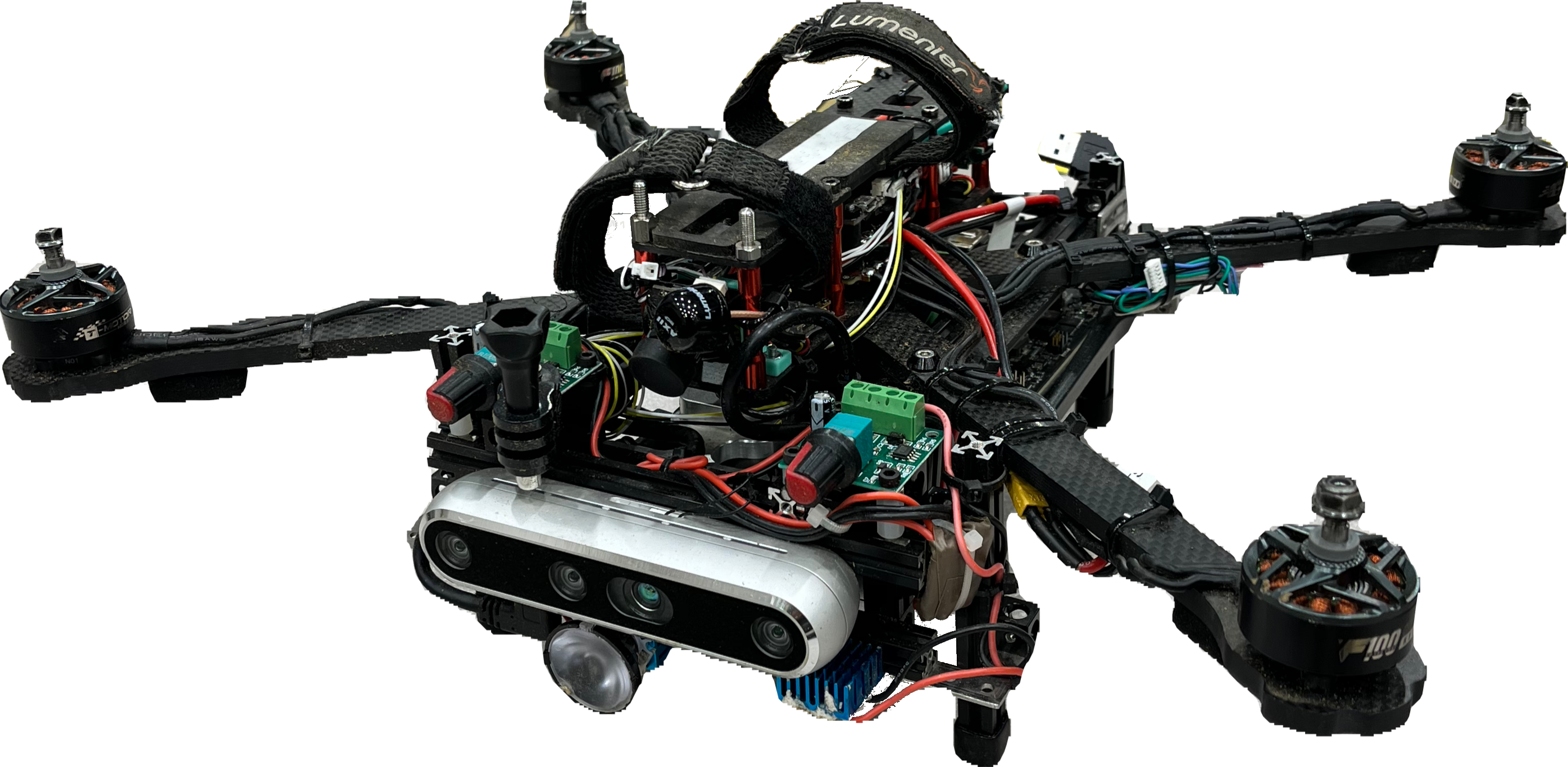}
    \includegraphics[width=0.39\linewidth]{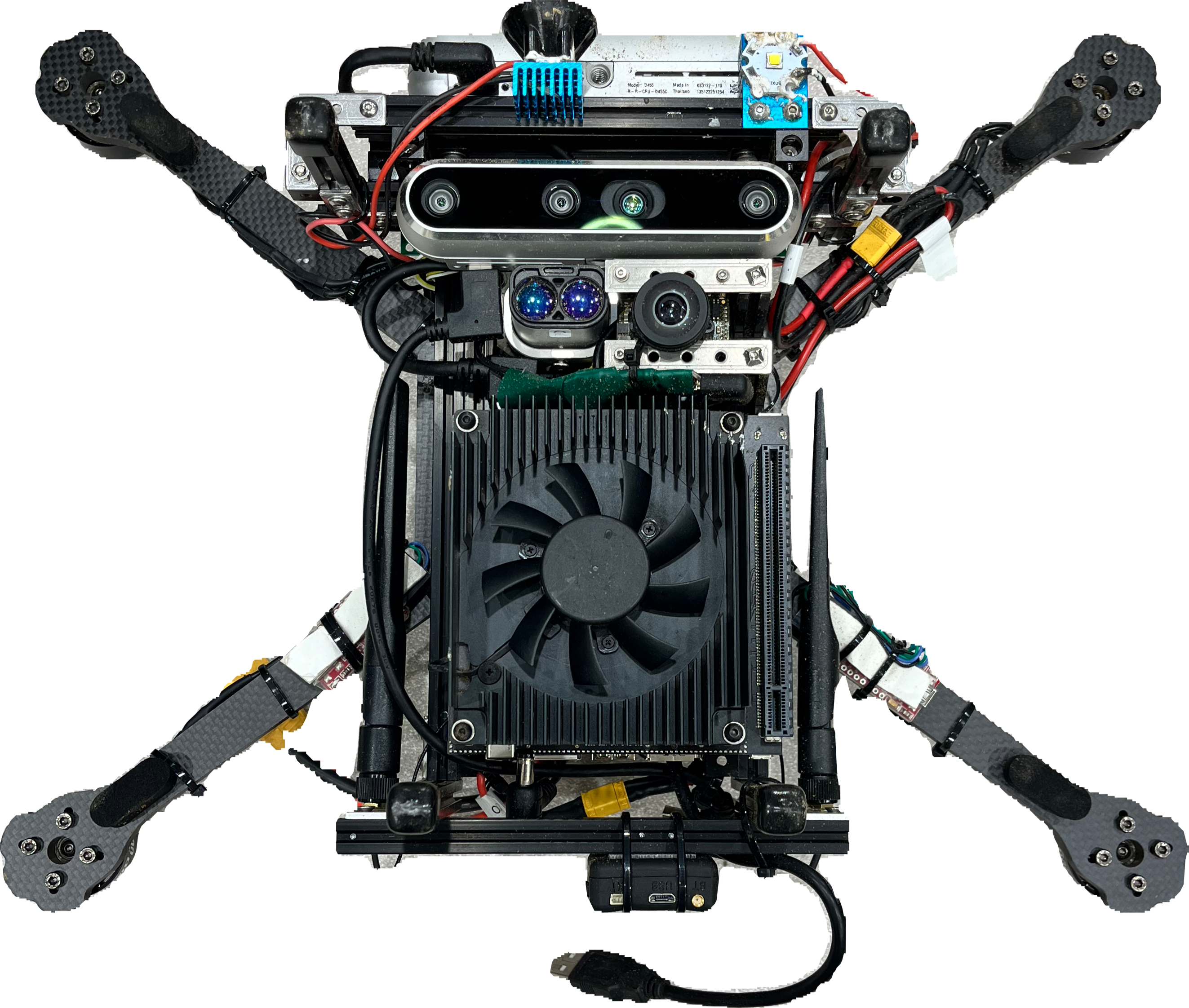}
    \caption{The aerial robot used in hardware experiments is
        equipped with a D455 and NVIDIA Orin AGX.
    \label{fig:omicron02}}
\end{figure}

A custom quadrotor platform is developed for evaluation in
hardware experiments. It is equipped with a forward-facing
Intel RealSense D455~(\cref{fig:omicron02}), a downward-facing
LW20 range finder, and a Matrix Vision BlueFox global
shutter greyscale camera. The moncular visual inertial
navigation system of~\citet{Yao-2020-120702} is used for GPS-denied
state estimation. The system is fully autonomous and all
calculations are conducted onboard a NVIDIA Orin AGX equipped with 32
GB of RAM. The total system mass is \SI{2.8}{\kilogram}.
The RealSense generate depth images at 30 Hz with a
resolution of 424 $\times$ 240. The re-planning algorithm
runs at \SI{12}{\Hz}, and
maintain a \SI{1}{\second} history of frames. We utilize a cascaded
controller (the outer loop runs at
\SI{100}{\hertz}), which runs on the NVIDIA Orin. An mRo
Pixracer flight controller runs the attitude control loop at a higher
frequency with modified PX4 firmware.

Experiments are conducted in three environments: (1) an outdoor
flight arena, (2) forest, and (3) cave.

\begin{figure*}[h]
    \centering
    \includegraphics[width=0.49\linewidth]{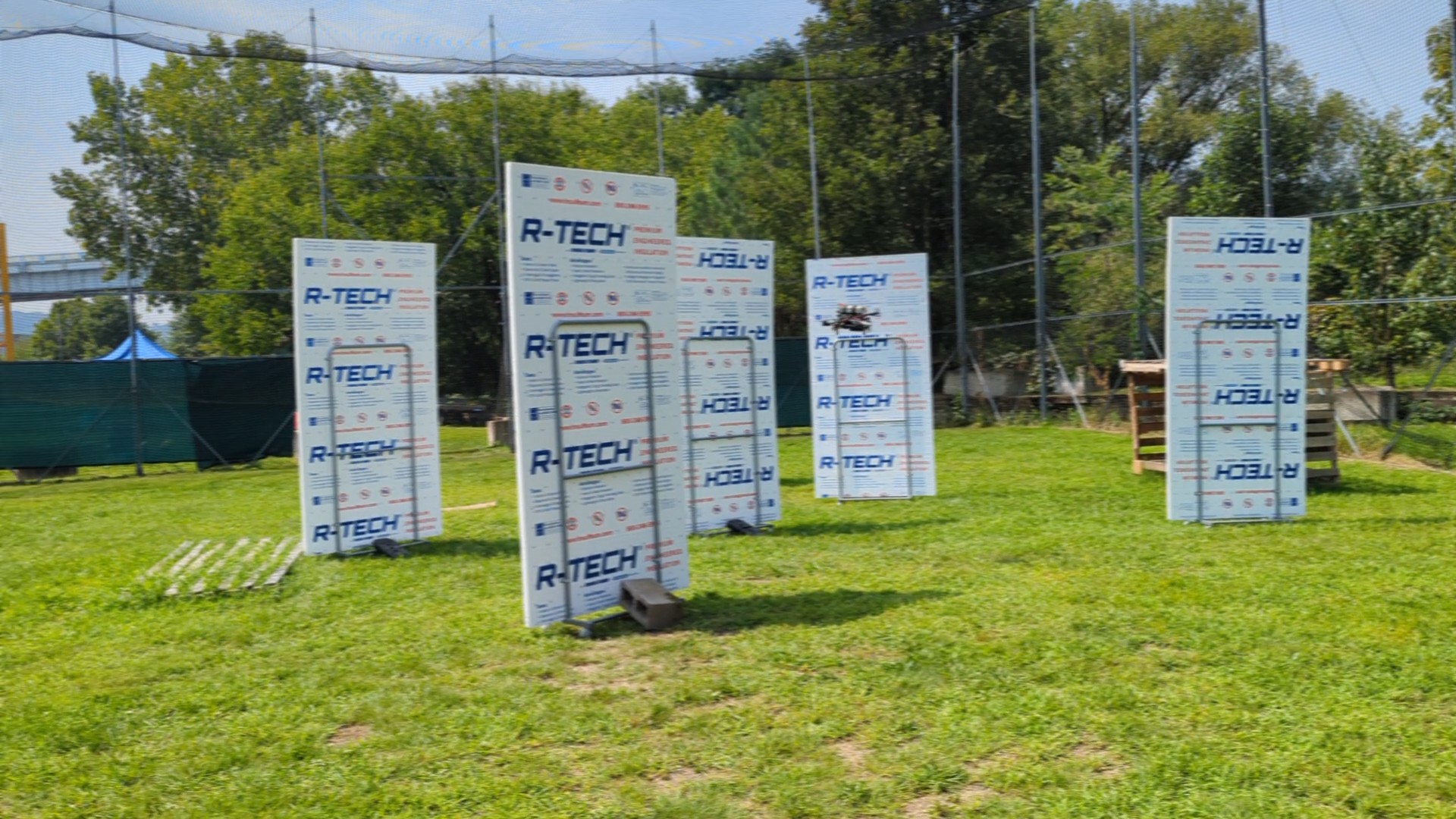}
    \includegraphics[width=0.49\linewidth]{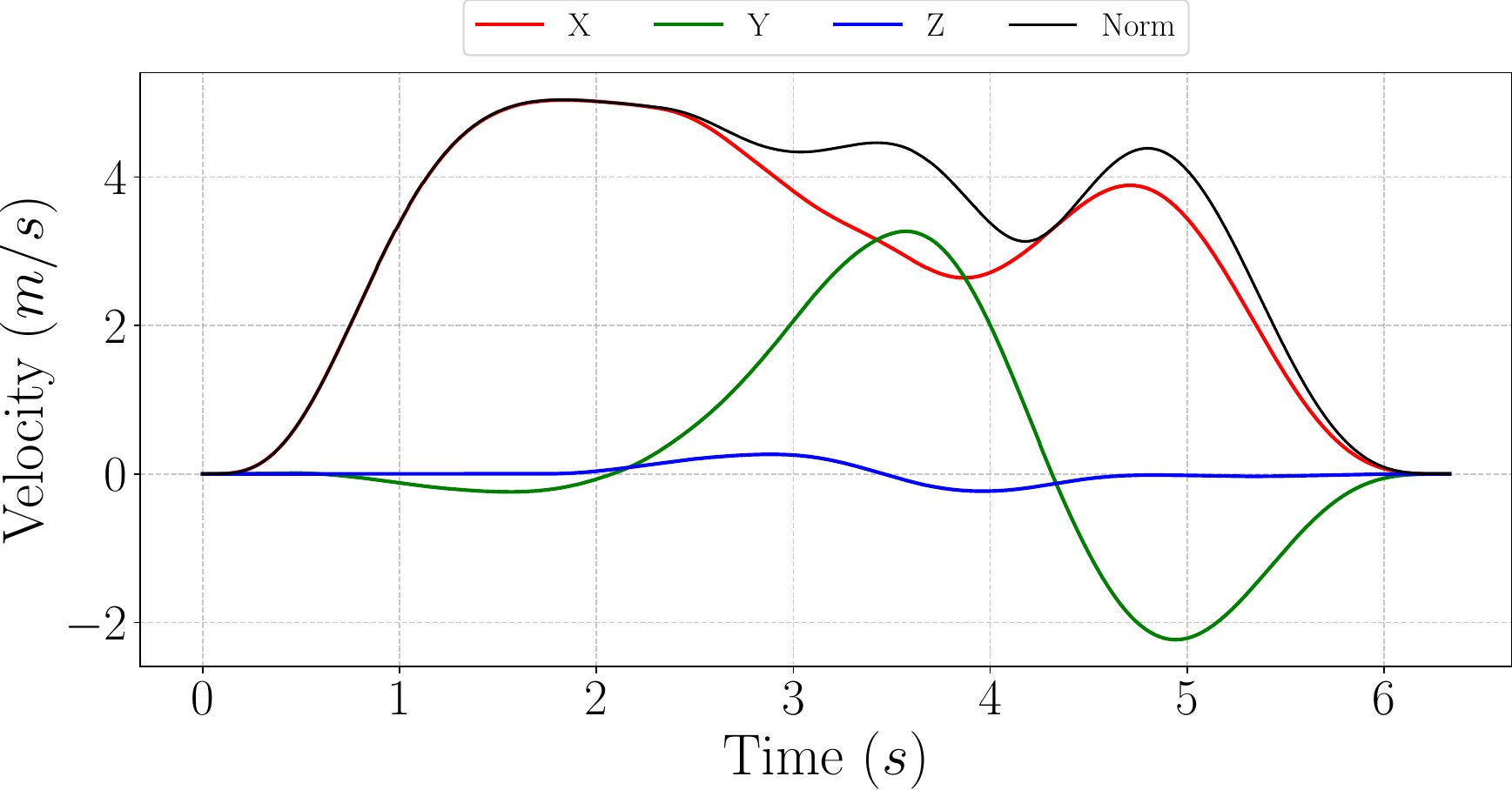}
    \includegraphics[width=0.49\linewidth]{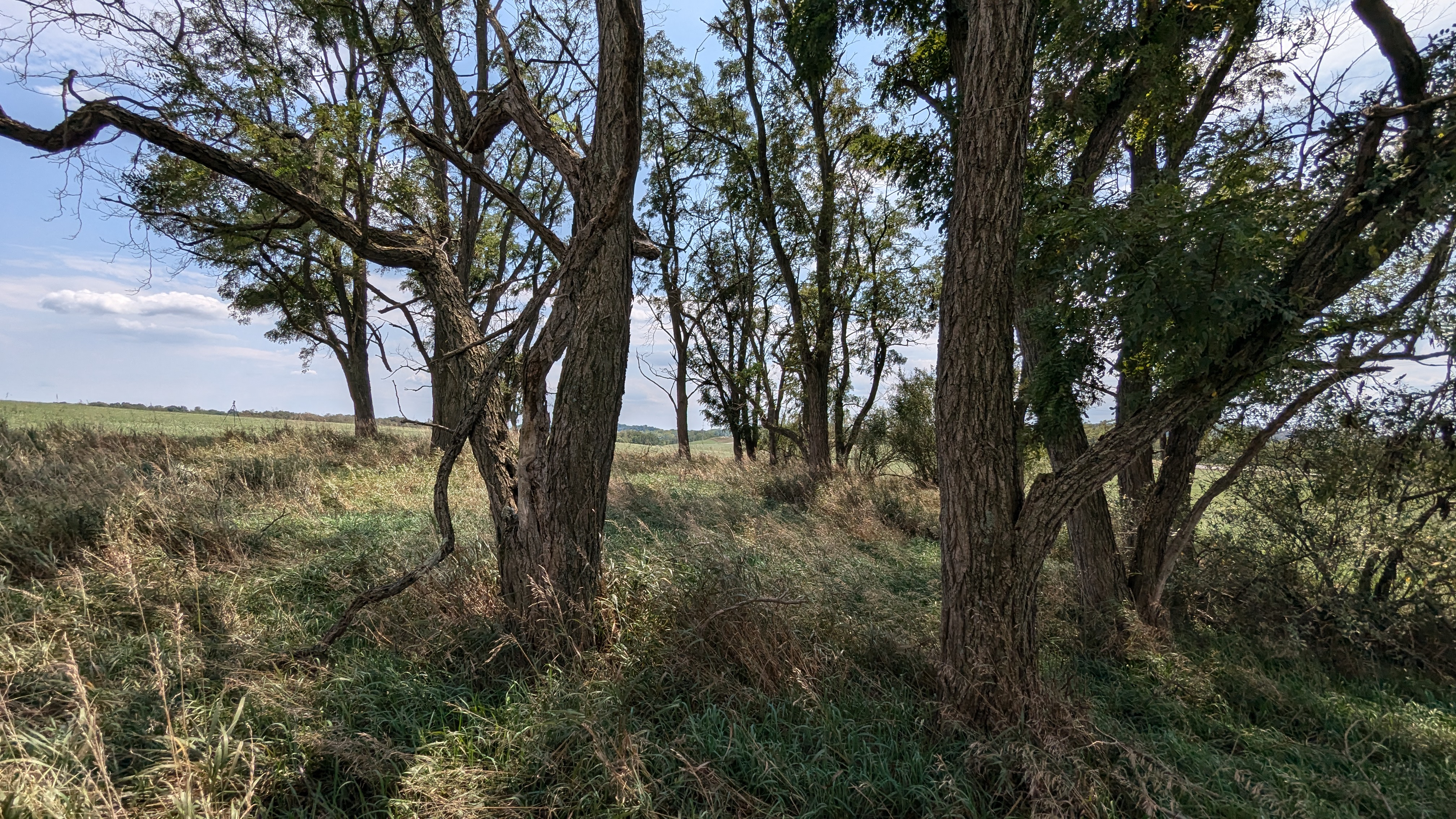}
    \includegraphics[width=0.49\linewidth]{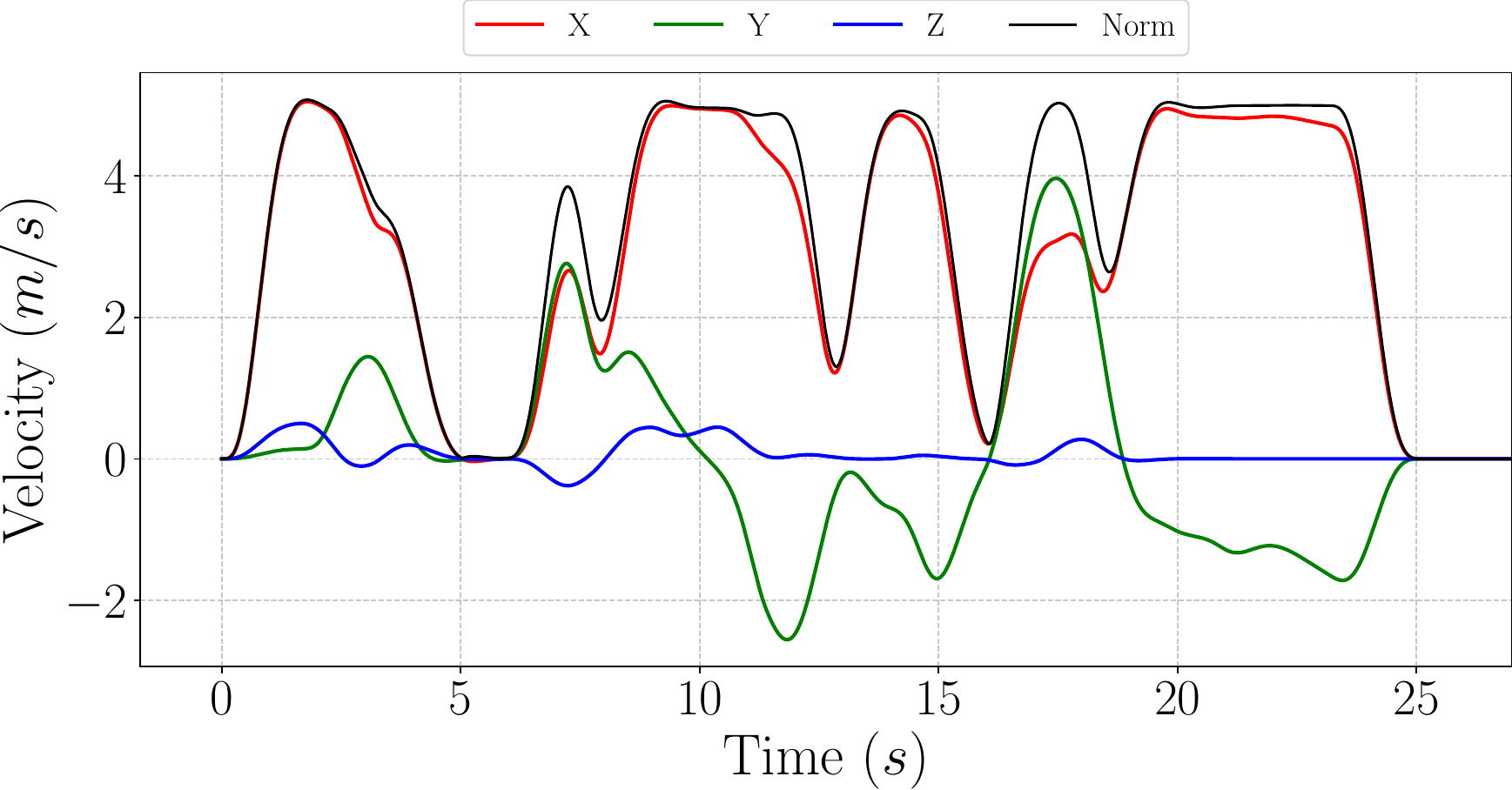}
    \includegraphics[width=0.49\linewidth]{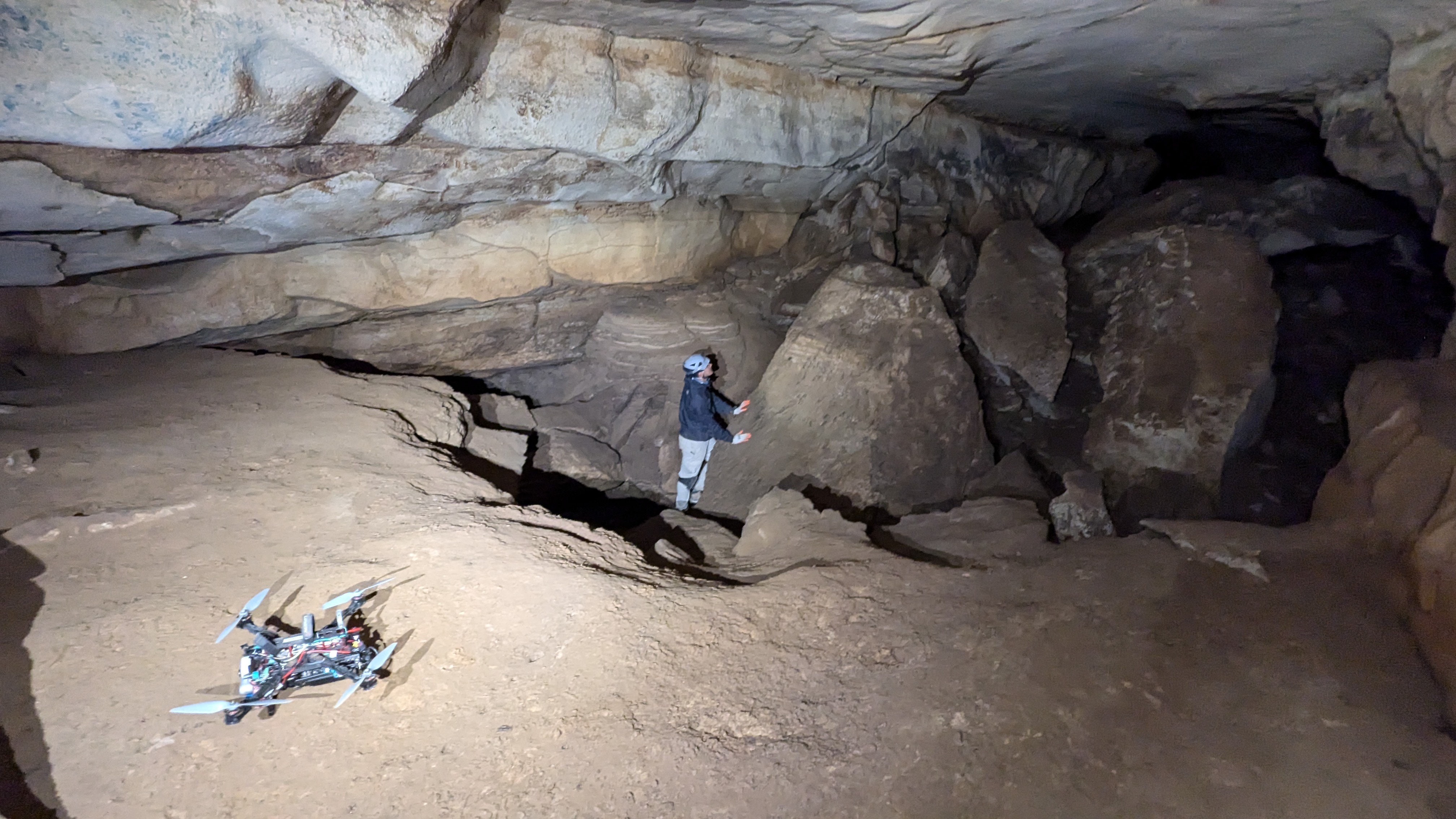}
    \includegraphics[width=0.49\linewidth]{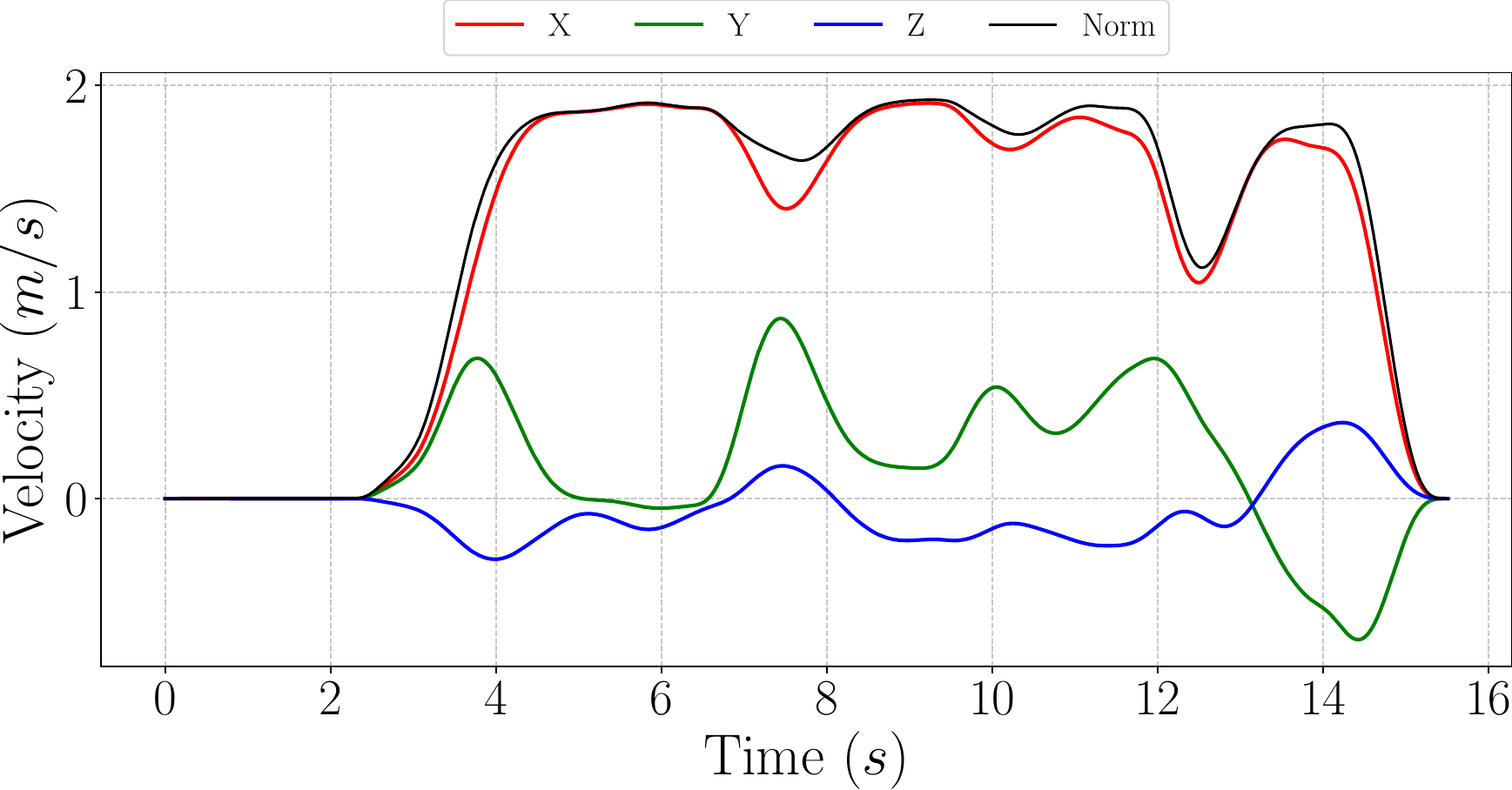}
    \caption{Left: flight arena, forest, and cave test environments. Right:
      reference trajectory velocity profiles for flights 10, 7 and 4, respectively, from~\cref{tab:hardware}. The maximum
        achieved speed across all trials is \SI{6}{\meter\per\second}.
    \label{fig:scenes_and_velocities}}
\end{figure*}

\begin{figure*}
  \includegraphics[width=0.999\linewidth]{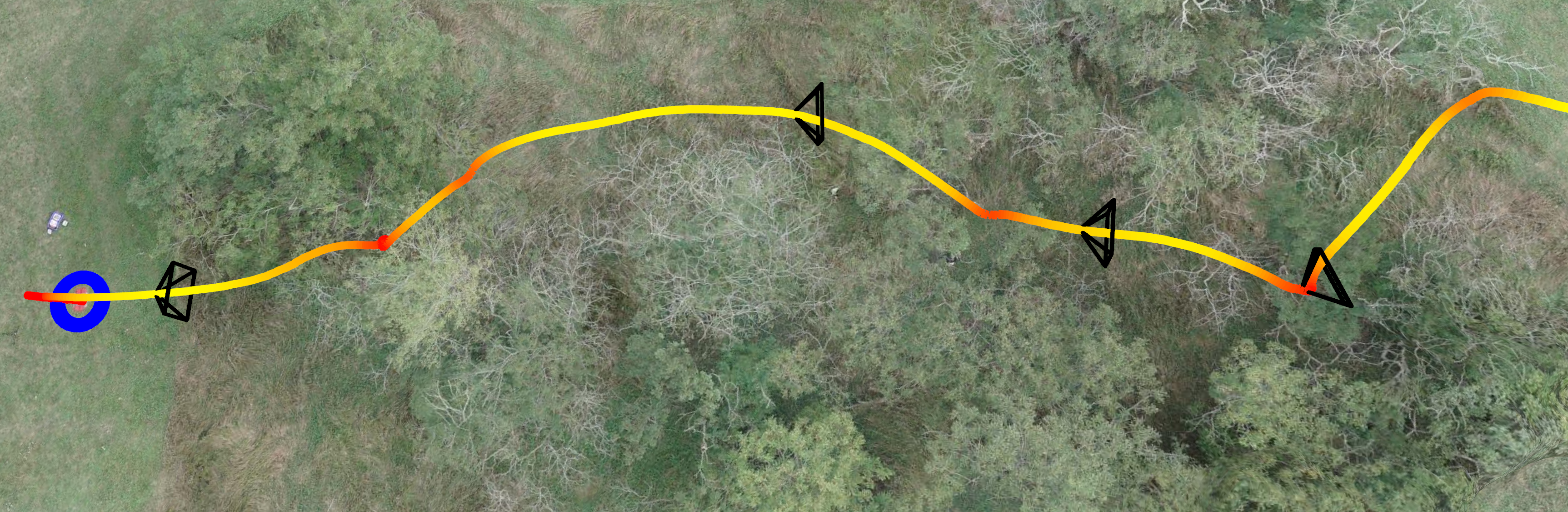}%
  \vspace{0.1cm}
  \includegraphics[width=0.245\linewidth,trim=0 1200 2300 160,clip]{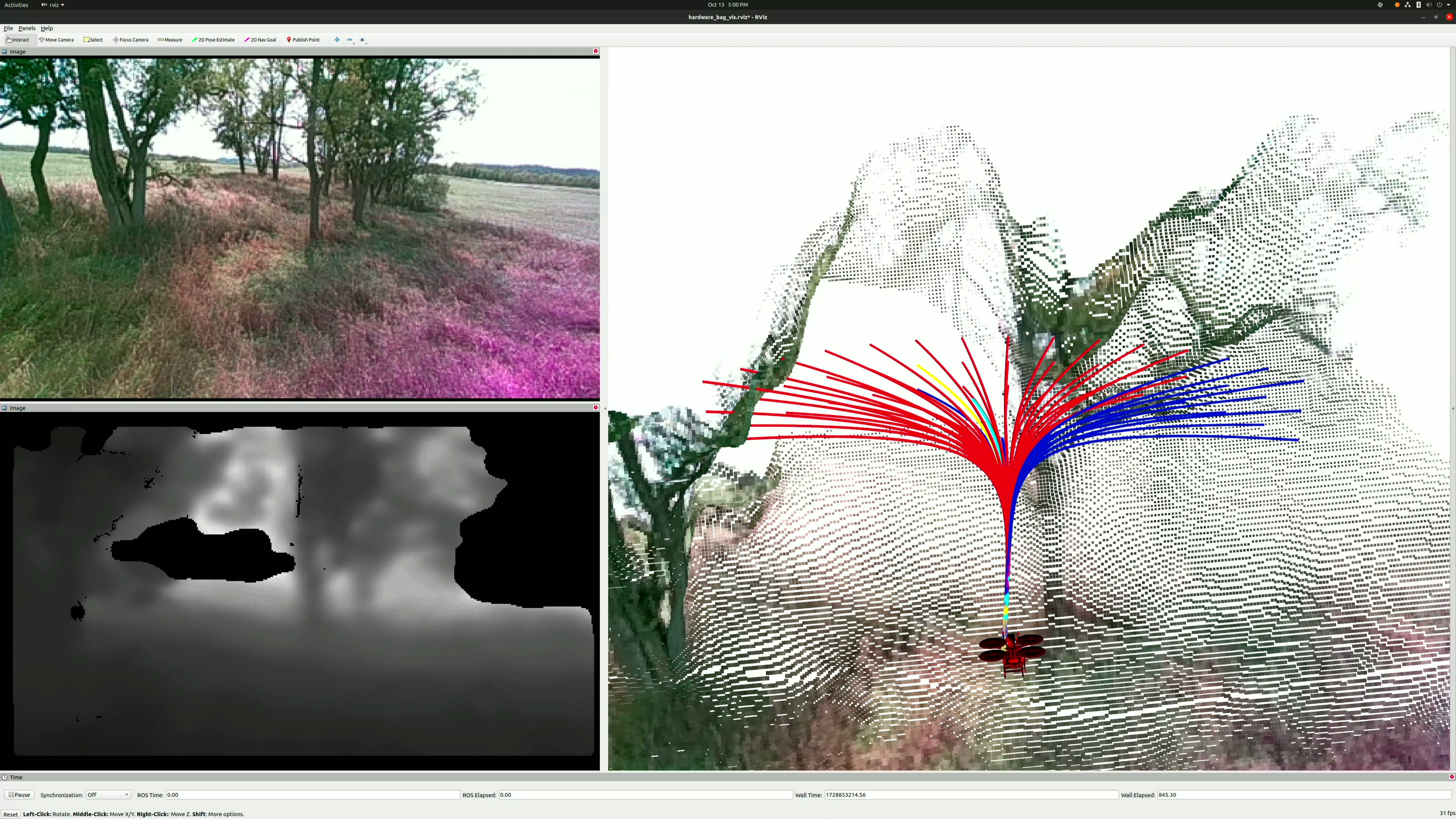}\hfill
  \includegraphics[width=0.245\linewidth,trim=0 1200 2300 160,clip]{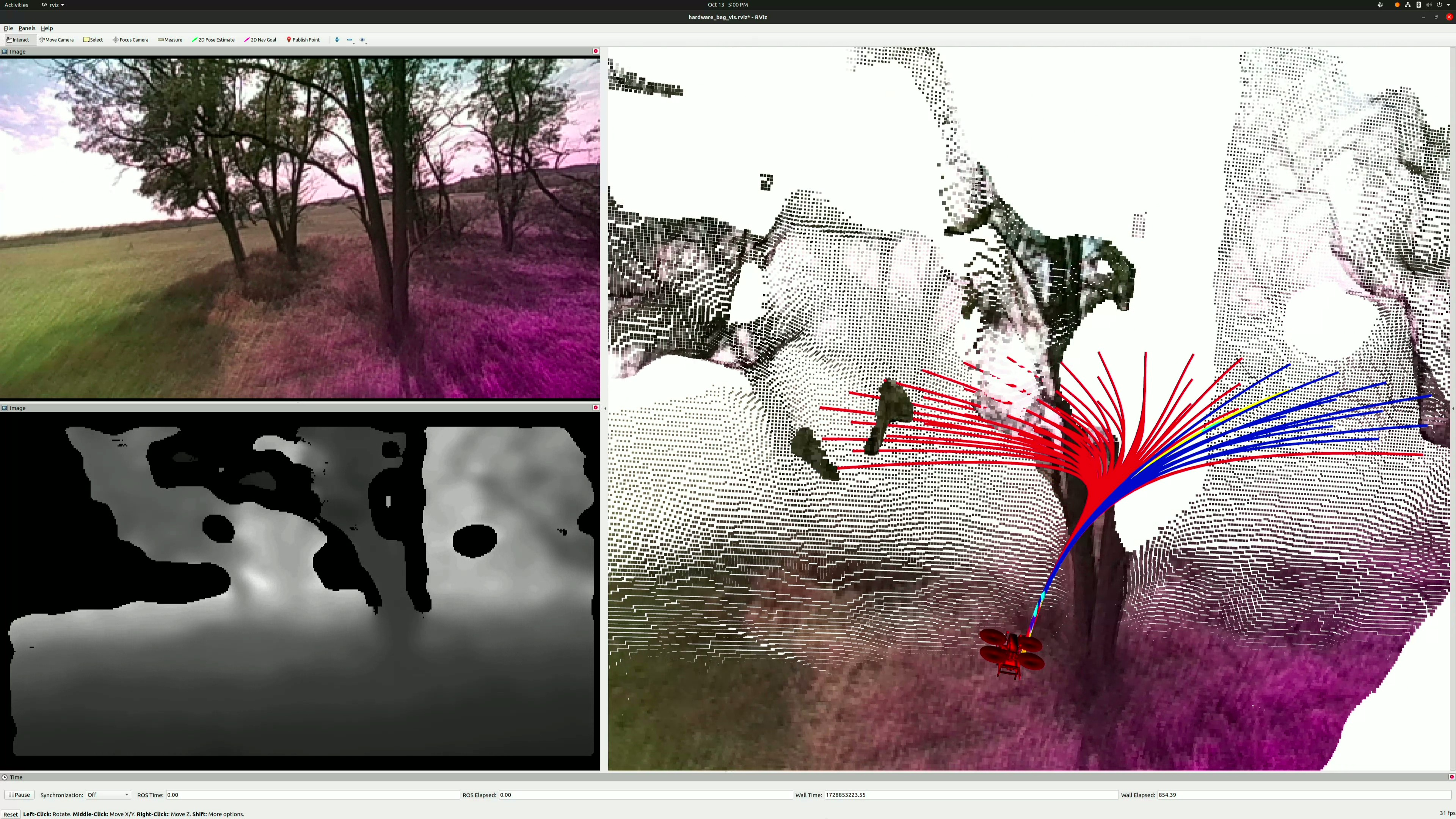}\hfill
  \includegraphics[width=0.245\linewidth,trim=0 1200 2300 160,clip]{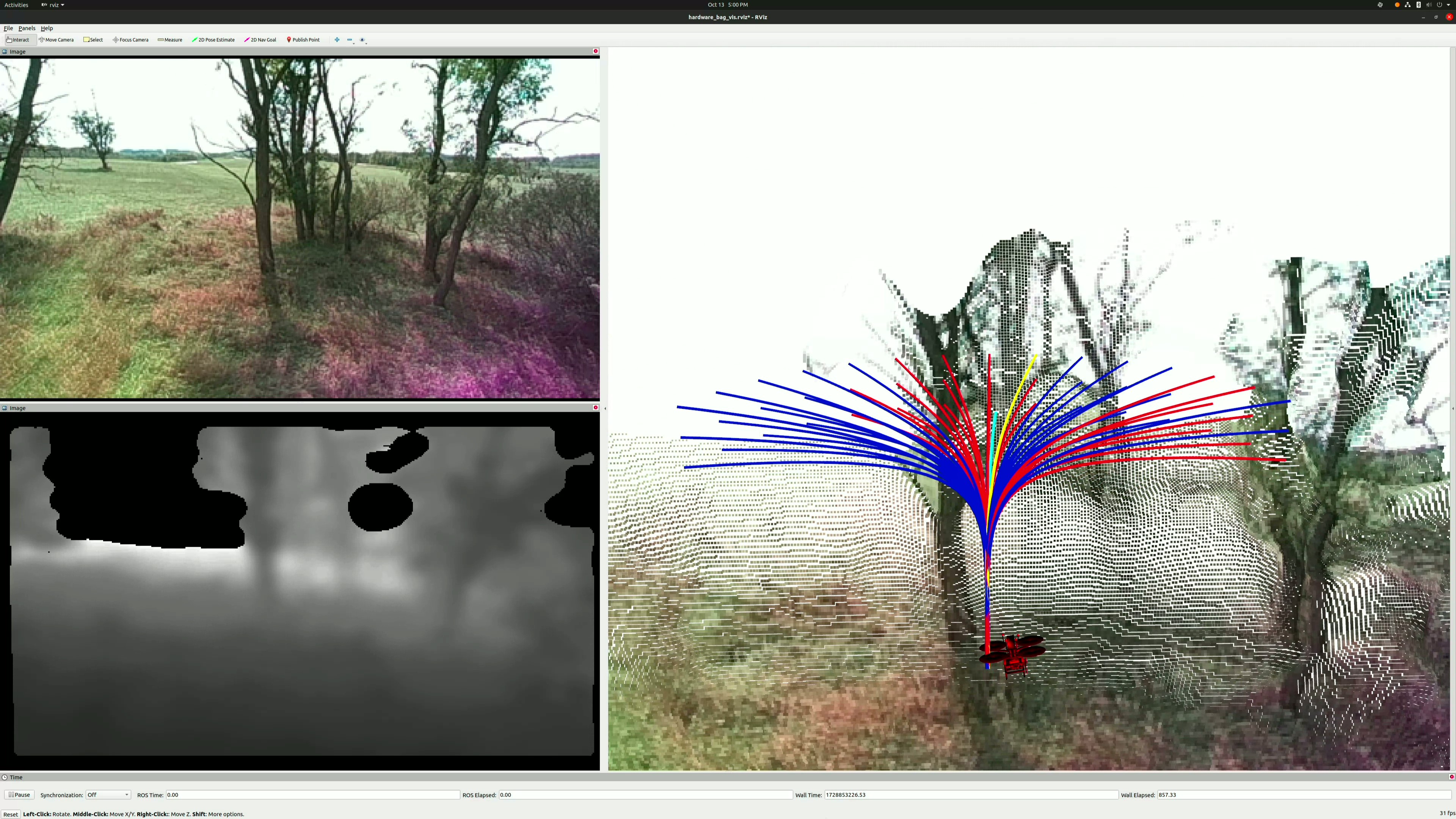}\hfill
  \includegraphics[width=0.245\linewidth,trim=0 1200 2300 160,clip]{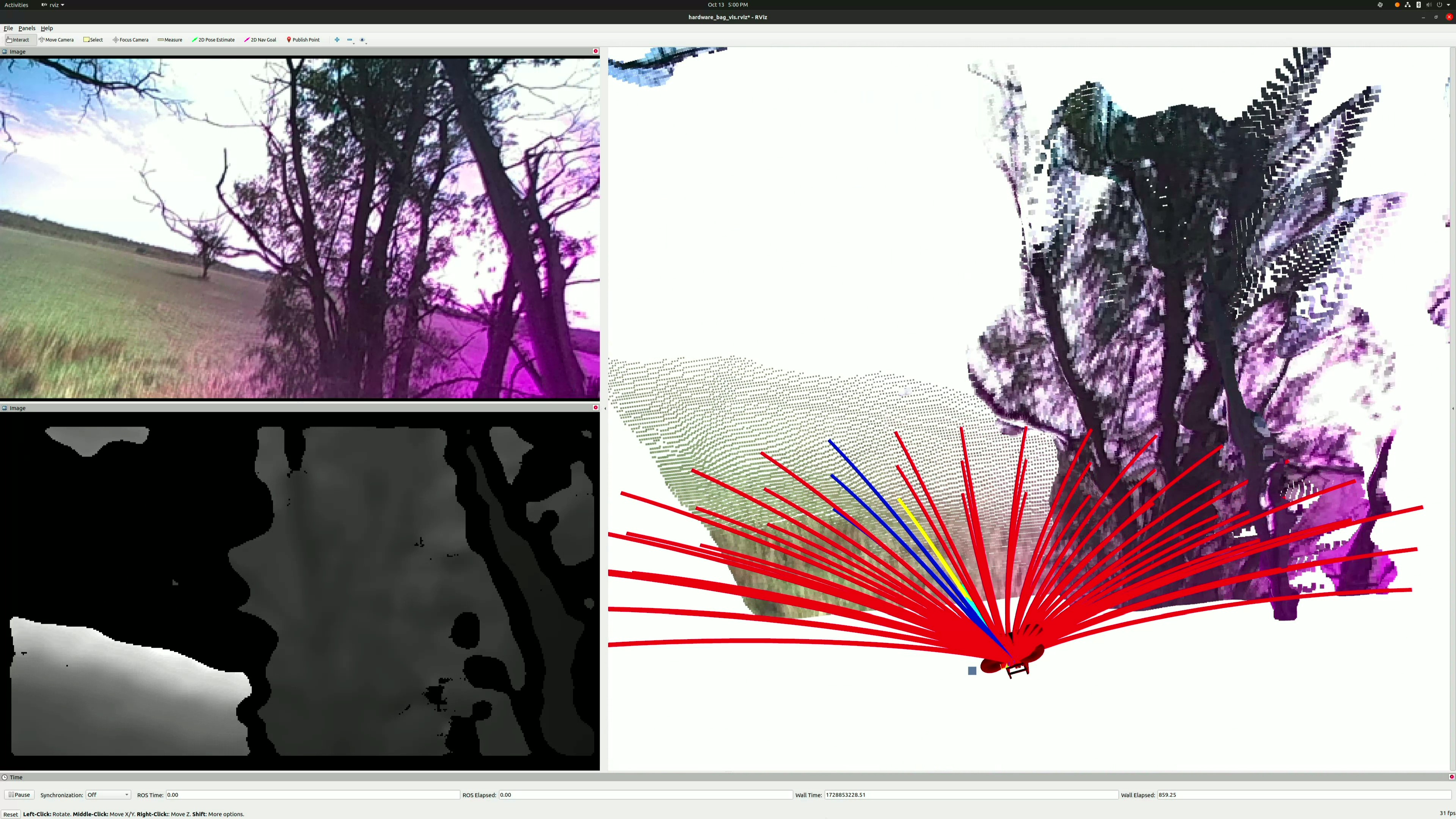}%
  \vspace{0.1cm}
  \includegraphics[width=0.245\linewidth,trim=1700 250 20 700,clip]{figures/hardware/240828_forest/screenshots_with_mpl/mplframe01.jpg}\hfill
  \includegraphics[width=0.245\linewidth,trim=1700 250 20 700,clip]{figures/hardware/240828_forest/screenshots_with_mpl/mplframe02.jpg}\hfill
  \includegraphics[width=0.245\linewidth,trim=1700 250 20 700,clip]{figures/hardware/240828_forest/screenshots_with_mpl/mplframe04.jpg}\hfill
  \includegraphics[width=0.245\linewidth,trim=1700 250 20 700,clip]{figures/hardware/240828_forest/screenshots_with_mpl/mplframe05.jpg}
  \caption[]{Outdoor obstacle avoidance test under
    tree canopy (forest flight 7 in \cref{fig:scenes_and_velocities} and
    \cref{tab:hardware}). Top: VINS trajectory overlayed on terrain map (low speeds are shown in red and high speeds in yellow). Camera
    frustums are visualized at four time steps where the robot maneuvers to
    avoid obstacles. Middle: Onboard RGB image from RealSense D455. Bottom:
    Robot and forward-arc motion primitive library (red: non-safe primitives,
    blue: safe primitives, yellow: selected primitive, cyan: stopping
    primitive).\label{fig:forest_flight}}%
  \vspace{-0.2cm}
\end{figure*}

\begin{table}[h!]
\centering
\caption{Hardware Flight Trials\label{tab:hardware}}
\resizebox{\columnwidth}{!}{
\begin{tabular}{llrrrrr}
\toprule
Env. & Flight & Flight Time & Path Length & $v_{max}$ & Max Speed & Avg Speed \\
& $\#$ & $s$ & $m$ & $m/s$ & $m/s$ & $m/s$ \\
\midrule
\multirow{11}{*}{\rotatebox{90}{Outdoor Flight Arena}}
    & 1 & 12.08 & 16.45 & 2 & 2.21 & 1.32 \\
    & 2 & 9.62 & 17.48 & 3 & 3.04 & 1.76 \\
    & 3 & 7.01 & 17.89 & 4 & 4.09 & 2.29 \\
    & 4 & 10.71 & 17.69 & 3 & 3.04 & 1.52 \\
    & 5 & 9.40 & 18.71 & 4 & 4.12 & 1.83 \\
    & 6 & 8.12 & 17.58 & 4 & 4.01 & 2.06 \\
    & 7 & 6.94 & 17.85 & 5 & 4.85 & 2.46 \\
    & 8 & 7.65 & 19.85 & 5 & 5.11 & 2.43 \\
    & 9 & 9.22 & 20.49 & 5 & 5.09 & 2.11 \\
    & 10 & 8.36 & 22.14 & 5 & 5.02 & 2.54 \\
    & 11 & 7.13 & 18.23 & 6 & 6.06 & 2.38 \\
\midrule
\multirow{9}{*}{\rotatebox{90}{Forest}}
    & 1 & 9.73 & 19.88 & 3 & 3.11 & 1.90 \\
    & 2 & 9.68 & 14.77 & 3 & 3.50 & 1.42 \\
    & 3$^\clubsuit$ & 11.57 & 29.91 & 3 & 3.30 & 2.53 \\
    & 4 & 9.15 & 30.36 & 5 & 5.13 & 3.10 \\
    & 5 & 18.43 & 64.20 & 5 & 5.20 & 4.34 \\
    & 6$^\clubsuit$ & 4.77 & 15.86 & 5 & 5.07 & 3.25 \\
    & 7 & 33.62 & 91.28 & 5 & 5.22 & 2.58 \\
    & 8 & 22.23 & 87.79 & 5 & 5.29 & 3.81 \\
    & 9$^\clubsuit$ & 36.40 & 29.47 & 5 & 5.05 & 0.71 \\
\midrule
\multirow{6}{*}{\rotatebox{90}{Cave}}
    & 1 & 7.79 & 10.30 & 2 & 1.91 & 1.23 \\
    & 2 & 13.98 & 14.23 & 2 & 2.08 & 0.86 \\
    & 3 & 11.05 & 12.98 & 2 & 2.05 & 1.11 \\
    & 4 & 13.96 & 20.97 & 2 & 2.34 & 1.39 \\
    & 5$^\blacklozenge$ & 11.69 & 13.45 & 3 & 3.43 & 1.06 \\
    & 6$^\spadesuit$ & 6.05 & 12.04 & 3 & 3.16 & 1.87 \\
\bottomrule
\multicolumn{7}{l}{\footnotesize $^\clubsuit$ indicates a crash due to thin
obstacle not being detected.}\\
\multicolumn{7}{l}{\footnotesize $^\blacklozenge$ indicates robot did not find a safe route and timed out.}\\
\multicolumn{7}{l}{\footnotesize $^\spadesuit$ indicates robot crashed due to VINS state estimate drift.}\\
\end{tabular}
}
\vspace{1cm}
\end{table}

\subsubsection{Outdoor Drone Arena}
11 high speed flight experiments were conducted in an outdoor
space with an area of \SI{18}{\meter} $\times$
\SI{24}{\meter}. Foam boards with dimension \SI{2}{\meter}
high by \SI{1}{\meter} wide are used as obstacles.
The flight statistics for the 11 flight experiments are shown
in~\cref{fig:scenes_and_velocities,tab:hardware}.  The goal location
for all trials was set to be \SI{18}{\meter} from the
starting point, while desired speed and obstacle locations varied over
the trials. The maximum speed achieved was 6 m/s. All 11 trials were
successful (over \SI{205}{\meter} total path length).

\subsubsection{Forest}
The approach was tested under the canopy of a stand of trees
over 9 trials. Start and goal positions varied with a goal
distance of \SI{15}{\meter} up to \SI{80}{\meter}. The maximum forward
velocity was set to \SI{5}{\meter\per\second}.
A trial is visualized in~\cref{fig:forest_flight}

\subsubsection{Cave}
The final set of experiments deployed the robot to a cave in Kentucky. The cave
features large rock outcrops and a deep chasm. The goal position varied between
\SI{10}{\meter} and \SI{20}{\meter} away. Over 6 trials, four trials succeeded,
one timed out, and one ended in collision. The cave features narrow sections
with a vertical gap less than \SI{1.8}{\meter}.  The desired forward velocity
was set between \SI{2}{\meter\per\second} and \SI{3.5}{\meter\per\second}.

\subsection{Discussion on failure modes} As shown in \cref{tab:hardware}
we experienced three failure modes in hardware experiments.  First, perception
errors due to thin branches that were not detected by the depth sensor led to 3
collisions in the forest environment.  Second, without a global planner, the
robot got stuck in a dead end in the cave environment where navigation to the
goal required backtracking.  Lastly, cave flight 6 experienced a collision
due to VINS state estimation drift. During execution of a stopping maneuver,
particulates from rotor downwash caused incorrect feature matches in the
downward-facing camera, which led to the robot tracking the stopping trajectory
into the ground instead of at the correct altitude.

\section{Conclusion\label{sec:conclusion}}
In this work we propose a reactive navigation method using a history of depth
images and forward-arc motion primitives. We evaluate the performance of
\emph{Forward-Arc} against two baseline reactive methods in diverse simulation
environments.  \emph{Forward-Arc} achieves the highest success rate and lowest
collision rate in all environments, demonstrating the efficacy of the primitive
scheduling approach and safe stopping strategy.  Additionally,
\emph{Forward-Arc} reaches the objective in obstacle density experiments with
lower average flight time, path length, and control effort metrics compared to
the baselines.  Finally, we conduct hardware trials in a flight arena, forest, and
cave environments to validate the approach in diverse, real-world conditions.
A maximum speed of \SI{6}{\meter\per\second} is achieved over \SI{571}{\meter}
without collisions in successful trials.  Potential avenues for future work include
adapting the desired forward velocity to varying levels of environment clutter
and incorporating a hierarchical global planner to enable the robot to escape
dead ends.

\section{Acknowledgment}
The authors would like to thank E. Burkholder for field testing
support. The authors would also like to thank K. Bailey and
T. Miller for facilitating experiments at the cave in Kentucky.

\balance
\bibliographystyle{IEEEtranN}
{
  \footnotesize
  \bibliography{refs}
}

\end{document}